# Causal Feature Selection Framework for Stable Soft Sensor Modeling based on Time-Delayed Cross Mapping

Shi-Shun Chen [a, b], Xiao-Yang Li [a,*], Enrico Zio [b, c]

[a] School of Reliability and Systems Engineering, Beihang University, Beijing, China
[b] Energy Department, Politecnico di Milano, Milan, Italy
[c] Centre de Recherche sur les Risques et les Crises (CRC), MINES Paris-PSL University, Sophia Antipolis, France

Email:
css1107@buaa.edu.cn (Shi-Shun Chen)
leexy@buaa.edu.cn (Xiao-Yang Li)
enrico.zio@polimi.it, enrico.zio@mines-paristech.fr (Enrico Zio)

Corresponding author[*]: Xiao-Yang Li

# Highlights

- Varying causal strength across time delay is considered for feature selection.
- State space reconstruction is employed to deal with interdependent variables.
- Time-delayed partial cross mapping is proposed for direct causal inference.
- Features are selected automatically via causal strength and validation performance.
- Features selected by direct causality can enhance model stability.

# Abstract

Soft sensor modeling plays a crucial role in process monitoring. Causal feature selection can enhance the performance of soft sensor models in industrial applications. However, existing methods ignore two critical characteristics of industrial processes. Firstly, causal relationships between variables always involve time delays, whereas most causal feature selection methods investigate causal relationships in a same time dimension. Secondly, variables in industrial processes are often interdependent, which contradicts the decorrelation assumption of traditional causal inference methods. Consequently, soft sensor models based on existing causal feature selection approaches often lack sufficient accuracy and stability. To overcome these challenges, this paper proposes a causal feature selection framework based on time-delayed cross mapping. Time-delayed cross mapping employs state space reconstruction to effectively handle interdependent variables in causality analysis, and considers varying causal strength across time delay. Time-delayed convergent cross mapping (TDCCM) is introduced for total causal inference, and time-delayed partial cross mapping (TDPCM) is developed for direct causal inference. Then, in order to achieve automatic feature selection, an objective feature selection strategy is presented. The causal threshold is automatically determined based on the model performance on the validation set, and the causal features are then selected. Two real-world case studies show that TDCCM achieves the highest average performance, while TDPCM improves soft sensor stability and performance in the worst scenario. On average over the two cases, TDCCM decreases root mean square error (RMSE) by about 8.93% compared with the best existing feature selection method, and TDPCM further decreases RMSE in the worst scenario by about 7.69% relative to TDCCM. The code is publicly available at https://github.com/dirge1/TDPCM.

# List of Symbols

| | |
|---|---|
| **C** | Candidate set of causal thresholds for feature selection |
| $D$ | Number of candidates for causal threshold determination |
| $E$ | Embedding dimension |
| $L$ | Length of the time series |
| $M$ | Number of auxiliary variables for soft sensor modeling |
| $N$ | Number of historical measurements for soft sensor modeling |
| $\mathbf{M}_X, \mathbf{M}_Y, \mathbf{M}_Z$ | Trajectory matrices of $X$, $Y$ and $Z$ |
| $\mathbf{M}_{X,l}, \mathbf{M}_{Y,l}$ | Embedding vectors of $X$ and $Y$ in the reconstructed space at the time point $l$ |
| $\hat{\mathbf{M}}_{Y,l}^{X \to Y}$ | Predicted embedding vectors of $Y$ at the time point $l$ using the trajectory matrix of $X$ |
| $\hat{\mathbf{M}}_{Y,l-\xi}^{X \to Y, \xi}$ | Predicted embedding vectors of $Y$ at the time point $l - \xi$ using the trajectory matrix of $X$ considering time delay $\xi$ |
| $Q$ | Number of disturbing variables |
| $\mathbf{U}_\gamma$ | Stacked vector at time delay $\gamma$ for TDPCM calculation |
| $X, Y, Z$ | Time series variables |
| $\mathbf{Z}$ | Disturbing variable set |
| $\hat{Z}_X$ | Predicted time series of $Z$ using the trajectory matrix of $X$ |
| $c_1, c_2, \ldots, c_D$ | Candidates for causal threshold determination |
| $c_{\text{best}}$ | Optimal causal threshold for feature selection |
| $d$ | Maximum time delay considered in TDCCM/TDPCM analysis |
| $f$ | Mapping function of the soft sensor model |
| $l$ | Generic notation for time point |
| $t_{\min}$ | First valid prediction time for reconstructed series |
| $t_i^l$ | Time point of the $i^{\text{th}}$ nearest neighbor embedding vector |
| $w_i$ | Weight assigned to the $i^{\text{th}}$ nearest neighbor based on Euclidean distance |
| $x(l), y(l)$ | The value of $X$ and $Y$ at the time point $l$ |
| $\mathbf{y}_{t_{\min}:L}$ | Time series of $Y$ from time point $t_{\min}$ to time point $L$ |
| $\hat{y}_{X \to Y}(l)$ | Scale prediction of $y(l)$ using the trajectory matrix of $X$ |
| $\hat{\mathbf{y}}_{t_{\min}:L}^{X \to Y}$ | Predicted time series of $Y$ from time point $t_{\min}$ to time point $L$ using the trajectory matrix of $X$ |
| $\hat{y}_{X \to Y, \xi}(l - \xi)$ | Scale prediction of $y(l - \xi)$ using the trajectory matrix of $X$ considering time delay $\xi$ |
| $\hat{\mathbf{y}}_{t_{\min}:L-\xi}^{X \to Y, \xi}$ | Predicted time series of $Y$ from time point $t_{\min}$ to time point $L - \xi$ using the trajectory matrix of $X$ considering time delay $\xi$ |
| $\hat{\mathbf{y}}_{t_{\min}:L-\xi_{ZX}-\xi_{Y\hat{Z}_X}}^{X \to Z \to Y; \xi_{ZX}, \xi_{Y\hat{Z}_X}}$ | Predicted time series of $Y$ from time point $t_{\min}$ to time point $L - \xi_{ZX} - \xi_{Y\hat{Z}_X}$ using the predicted trajectory matrix of $Z$ considering time delay $\xi_{Y\hat{Z}_X}$, the predicted trajectory matrix of $Z$ is obtained using the trajectory matrix of $X$ considering time delay $\xi_{ZX}$ |
| $\mathbf{\Sigma}_\gamma$ | Covariance matrix of the stacked vector $\mathbf{U}_\gamma$ |
| $\mathbf{\Omega}_\gamma$ | Precision matrix corresponding to $\mathbf{\Sigma}_\gamma$ |
| $\mathbf{\Xi}_{YX}$ | Set of local maxima in the TDCCM causal strength curve from $Y$ to $X$ |
| $\mathbf{\Pi}_{YX}$ | Set of local maxima in the TDPCM causal strength curve from $Y$ to $X$ |
| $\gamma$ | Time delay considered in TDPCM for evaluating direct causal strength |
| $\gamma_{YX}$ | Optimal causal time delay from $Y$ to $X$ determined by TDPCM |
| $\xi$ | Generic notation for time delay |
| $\xi_{YX}$ | Optimal causal time delay from $Y$ to $X$ determined by TDCCM |
| $\rho_{Y \to X}$ | Causal strength from $Y$ to $X$ computed by CCM |
| $\rho_{Y \to X, \xi}$ | Time-delayed causal strength from $Y$ to $X$ at delay $\xi$ calculated by TDCCM |
| $\rho_{Y \to X|Z}$ | Direct causal strength from $Y$ to $X$ considering $Z$ calculated by PCM |
| $\rho_{Y \to X|Z, \gamma}$ | Time-delayed direct causal strength from $Y$ to $X$ considering $Z$ at delay $\gamma$ calculated by TDPCM |
| $\rho_{Y \to X|\mathbf{Z}, \gamma}$ | Time-delayed direct causal strength from $Y$ to $X$ considering the set $\mathbf{Z}$ at delay $\gamma$ calculated by TDPCM |
| $\tau$ | Embedding time delay |

## 1 Introduction

Monitoring, evaluating and optimizing industrial processes are crucial tasks. Generally, key performance indicators (KPIs) such as product quality, energy consumption and pollutant emissions are recorded continuously to reflect the state of the industrial process and provide guidance for process control. However, as industrial processes become increasingly complex, the cost and difficulty of direct KPI measurement have risen, making it challenging for online monitoring KPIs [1]. Thanks to the rapid development of the industrial internet of things, collecting abundant sensor data of easily measurable auxiliary variables have become feasible. By establishing mathematical relationships between auxiliary variables and KPIs, online KPI prediction can be achieved [2]. This technique is known as soft sensing (i.e., soft sensor modeling), and has been extensively adopted in practical industries [3].

There are two main approaches for developing soft sensor models: physics-based and data-driven [3]. The first approach is effective when the process physical mechanism is well-understood. Nevertheless, this prerequisite is always difficult to satisfy in actual industrial scenarios. Consequently, data-driven methods have emerged as vital alternatives. Representative data-driven approaches include statistical models like partial least square (PLS) regression [4, 5], and machine learning models like random forest (RF) regression [6]. Moreover, recently developed deep learning methods have also been applied to soft sensor modeling, such as long short-term memory (LSTM) neural network [7], convolutional neural network [8] and variational autoencoder [9]. Although data-driven methods have significantly boosted the development of soft sensor models, the stability of these models continues to be a major concern for their practical application.

Feature selection is one of the most effective strategies for ensuring the stability of soft sensor models, because it not only helps reveal the underlying process mechanism but also improves model robustness and reduces computational complexity. Existing feature selection approaches for industrial soft sensing mostly fall into three categories:

(1) **Correlation-based methods**, including techniques such as the Pearson correlation coefficient (PCC) [10] and grey relational analysis [11]. These methods identify auxiliary variables that are highly correlated with KPIs, and then select them as model inputs. However, correlation does not always imply causation [12]. Variables that significantly affect each other may exhibit weak correlations due to time delay, whereas highly correlated variables may lack causal relationship. Therefore, feature selection methods based on correlation may fall short in accurately predicting KPIs.

(2) **Model training-based approaches**, such as RF [13, 14], nonnegative garrote [15, 16] and attention mechanisms [17]. These methods automatically assign weights to input features during training. The weights help the model emphasize relevant features and minimize the influence of less significant ones. Nevertheless, these methods are highly dependent on the specific model architecture used, which can result in inconsistent assessments of feature importance across different models. Furthermore, these techniques may still capture correlations rather than uncovering causal relationships.

(3) **Causality-based methods**, including mutual information (MI) [18-21] and conditional mutual information (CMI) [22, 23], Granger causality (GC) [24, 25] and transfer entropy (TE) [26, 27], have been employed for feature selection in industrial processes. However, these methods exhibit several critical limitations when applied to industrial soft sensor modeling:

- Firstly, MI and CMI do not incorporate time delay into causal inference. They quantify causal influence between variables at the same time point, and therefore cannot capture causal relationships that manifest with time delays. This omission is problematic because industrial variables commonly exhibit delayed interactions due to control feedback loops, material transport or dynamic process responses [28, 29]. As a result, MI and CMI cannot identify causal features with meaningful time delays.

- Secondly, although GC and TE explicitly consider time delays in causal inference, they still suffer from two major limitations. On one hand, they rely on the decorrelation assumption, which requires that the influence of a causal variable on the target can be isolated by conditioning on other variables [30]. Real industrial processes violate this assumption because process variables are inherently interdependent [31]. Therefore, GC and TE may become unreliable in industrial processes. On the other hand, GC and TE cannot characterize how causal strength varies across different time delays. Both methods provide a single causal dependence result, making it impossible to determine the specific lag at which the causal effect is strongest or how the causal contribution evolves with increasing delay.

These issues highlight a gap between causal inference and practical feature selection in industrial processes and point to the need for a method that can quantify causal influence across multiple time delays and work reliably in interdependent industrial processes. To address this gap, this study proposes a causal feature selection framework tailored for industrial soft sensor modeling. The framework is supported by two time-delayed cross mapping methods: time-delayed convergent cross mapping (TDCCM) and time-delayed partial cross mapping (TDPCM), designed for inferring total and direct causality for interdependent time series across multiple time delays. Within these two methods, state-space reconstruction is employed to handle variable interdependence. By quantifying causal strength over lag dimensions, the framework enables the selection of informative variables and their time delays. Moreover, an objective feature selection strategy is developed, where the causal index threshold is automatically determined using validation performance rather than empirical rules.

The contributions of this work are summarized as follows:

- TDCCM and TDPCM are introduced for inference of total and direct causality between interdependent industrial variables across multiple time delays.
- A time-delayed causal feature selection framework tailored for industrial soft sensor modeling is proposed.
- An objective feature selection strategy based on the results of time-delayed causal inference techniques is presented involving causal threshold optimization.

Notably, the integration of causal inference techniques and graph neural networks is also a notable trend in the field of soft sensor model development [32-36]. This paper, however, focuses on feature selection using causal inference techniques, distinguishing it from the objective of graph neural networks, which is to further exploit spatial information from selected features based on causal graphs.

The organization of the paper is as follows. The preliminaries of convergent cross mapping and partial cross mapping are introduced in Section 2. Next, the time-delayed causal feature selection framework based on TDCCM and TDPCM is developed in Section 3. After that, the effectiveness of the proposed method in soft sensor modeling is verified by two engineering cases in Section 4. Subsequently, the effectiveness of the proposed causal inference method is demonstrated by numerical cases in Section 5. Finally, Section 6 concludes the work.

## 2 Preliminaries

### 2.1 Convergent Cross Mapping

The convergent cross mapping (CCM) method is grounded in the theory of state-space reconstruction, which stems from Takens' theorem [37]. This theorem posits that a time series can be embedded into a higher-dimensional space to reconstruct its dynamics. Considering two time series $X = \{x(t)\}_{t=1}^{L}$ and $Y = \{y(t)\}_{t=1}^{L}$ with length $L$, their state-space reconstruction is given by:

$$\begin{aligned} \mathbf{M}_{X,l} &= [x(l), x(l-\tau), x(l-2\tau), \ldots, x(l-(E-1)\tau)] \\ \mathbf{M}_{Y,l} &= [y(l), y(l-\tau), y(l-2\tau), \ldots, y(l-(E-1)\tau)] \end{aligned} \quad (1)$$

where $\mathbf{M}_{X,l}$ and $\mathbf{M}_{Y,l}$ are the embedding vectors of $X$ and $Y$ in the reconstructed space at the time point $l$; $\tau$ is the time delay of embedding; and $E$ is the embedding dimension. For the time series with length $L$, $l$ ranges from $1+(E-1)\tau$ to $L$, and totally $L-(E-1)\tau$ embedding vectors can be constructed from the time series. Then, the trajectory matrix of $X$ and $Y$ can be represented as:

$$\mathbf{M}_X = \begin{bmatrix} \mathbf{M}_{X,t_{min}} \\ \mathbf{M}_{X,t_{min}+1} \\ \cdots \\ \mathbf{M}_{X,L} \end{bmatrix}, \quad \mathbf{M}_Y = \begin{bmatrix} \mathbf{M}_{Y,t_{min}} \\ \mathbf{M}_{Y,t_{min}+1} \\ \cdots \\ \mathbf{M}_{Y,L} \end{bmatrix}, \quad t_{min} = 1 + (E-1)\tau \quad (2)$$

The trajectory matrix represents a set of sampled points constructed from monitored time series data. All the sampled points are on a manifold, which is a continuous structure in the reconstructed space that captures the time-varying states of the time series. According to the theory of CCM [30], if $X$ is causally influenced by $Y$, then the manifold of $X$ contains information that can reconstruct the dynamics of $Y$. At a specific time point $l$, the embedding vector of $Y$ can be predicted via a weighted approximation:

$$\hat{\mathbf{M}}_{Y,l}^{X \to Y} = \sum_{i=1}^{E+1} w_i \mathbf{M}_{Y,t_i^l} \quad (3)$$

where $w_i$ represents a weight determined by the distance between $\mathbf{M}_{X,l}$ and its $i^{th}$ nearest neighbor embedding vector with corresponding time point $t_i^l$; and $\mathbf{M}_{Y,t_i^l}$ represents the contemporaneous embedding vector of $Y$ at the time point $t_i^l$. The weights are calculated by [30]:

$$w_i = v_i / \sum_{j=1}^{E+1} v_j \quad (4)$$

$$v_i = \exp\left(-\frac{d\left(\mathbf{M}_{X,t_i^l}, \mathbf{M}_{X,l}\right)}{d\left(\mathbf{M}_{X,t_1^l}, \mathbf{M}_{X,l}\right)}\right) \quad (5)$$

where $d(\cdot,\cdot)$ is the Euclidean distance between two embedding vectors; and $\mathbf{M}_{X,t_1^l}$ represents the nearest embedding vector to $\mathbf{M}_{X,l}$ across all time points.

Subsequently, the scale prediction of $y(l)$ can be obtained by taking the first component of the predicted embedding vector:

$$\hat{y}_{X \to Y}(l) = \left(\hat{\mathbf{M}}_{Y,l}^{X \to Y}\right)_1 \quad (6)$$

By performing this reconstruction for all eligible time points, the predicted time series of $Y$ is expressed as:

$$\hat{\mathbf{y}}_{t_{min}:L}^{X \to Y} = \left[\hat{y}_{X \to Y}(t_{min}), \hat{y}_{X \to Y}(t_{min}+1), \ldots, \hat{y}_{X \to Y}(L)\right]^T \quad (7)$$

where $\hat{\mathbf{y}}_{t_{min}:L}^{X \to Y}$ represents the predicted time series of $Y$ from the time point $t_{min}$ to the time point $L$ using the trajectory matrix of $X$.

For clarity, we write the time series $X$ and $Y$ in vector form as:

$$\mathbf{x}_{1:L} = [x(1), x(2), \ldots, x(L)]^T, \quad \mathbf{y}_{1:L} = [y(1), y(2), \ldots, y(L)]^T \quad (8)$$

Then, the causal strength from $Y$ to $X$ is quantified by comparing the predicted and actual values through the Pearson correlation coefficient [30]:

$$\rho_{Y \to X} = \mathrm{pcc}\left(\hat{\mathbf{y}}_{t_{min}:L}^{X \to Y}, \mathbf{y}_{t_{min}:L}\right) \quad (9)$$

where pcc($\mathbf{e}_1$, $\mathbf{e}_2$) denotes the PCC calculation; and $\rho_{Y \to X}$ denotes the total causal strength from $Y$ to $X$ calculated by CCM. As the time series length $L$ approaches infinity, $\rho_{Y \to X}$ will converge to a specific value. If the convergent value exceeds the predefined threshold $c$, it means that $Y$ has a causal effect on $X$ and vice versa. Similarly, the dynamic of $X$ can be predicted based on the embedding vectors of $Y$ to detect the causal influence of *X* on *Y*.

2.2 Time-Delayed Convergent Cross Mapping

Suppose that there is a time delay $\xi$ in the causal relationship from *Y* to *X*. At this point, $\mathbf{M}_{x,l}$ is highly capable of predicting $y(l-\xi)$ theoretically [38], and the embedding vector of *Y* can be predicted as:

$$\hat{\mathbf{M}}_{Y,l-\xi}^{X \to Y,\xi} = \sum_{i=1}^{E+1} w_i \mathbf{M}_{Y,t_i^l} \qquad (10)$$

where the definitions of the parameters are consistent with those in Eq. (3). Owing to the time delay, there are $L-\xi-(E-1)\tau$ predictable time points for *Y*. The corresponding scalar prediction is obtained by taking the first component of the vector:

$$\hat{y}_{X \to Y,\xi}(l-\xi) = \left(\hat{\mathbf{M}}_{Y,l-\xi}^{X \to Y,\xi}\right)_1 \qquad (11)$$

Subsequently, after mapping the embedding vectors of *X* onto those of *Y* for all eligible time points, the predicted time series of *Y* is expressed as:

$$\hat{\mathbf{y}}_{t_{\min}:L-\xi}^{X \to Y,\xi} = \left[\hat{y}_{X \to Y,\xi}(t_{\min}), \hat{y}_{X \to Y,\xi}(t_{\min}+1), \ldots, \hat{y}_{X \to Y,\xi}(L-\xi)\right]^{\mathrm{T}} \qquad (12)$$

Then, by measuring the correlation between $\hat{\mathbf{y}}_{t_{\min}:L-\xi}^{X \to Y,\xi}$ and $\mathbf{y}_{t_{\min}:L-\xi}$, the causal strength of *Y* on *X* considering a time delay $\xi$ can be quantified. This is defined as time-delayed CCM (TDCCM) [38], which can be expressed as:

$$\rho_{Y \to X,\xi} = \mathrm{pcc}\left(\hat{\mathbf{y}}_{t_{\min}:L-\xi}^{X \to Y,\xi}, \mathbf{y}_{t_{\min}:L-\xi}\right) \qquad (13)$$

where $\rho_{Y \to X,\xi}$ denotes the total causal strength from *Y* to *X* at time delay $\xi$ calculated by TDCCM. When $\xi = 0$, TDCCM corresponds to CCM, and Eq. (13) is the same as Eq. (9).

2.3 Partial Cross Mapping

When the system has only two variables, their causal relationship is clearly direct. However, in a complex system with numerous variables, there can be two types of causation: direct and indirect, as illustrated in Fig. 1.

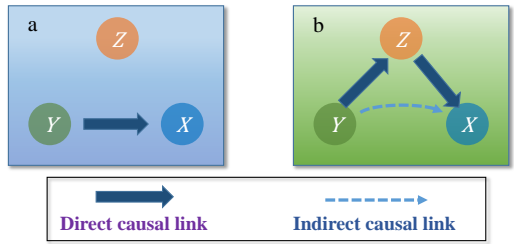

Fig. 1 Schematics of direct and indirect causal links. (a) *Y* directly causes *X*; (b) *Y* indirectly causes *X*.

To distinguish direct causations from indirect ones, Leng et al. [39] presented partial cross mapping (PCM) based on the framework of CCM. Considering a potential disturbing time series $Z = \{z(t)\}_{t=1}^{L}$, PCM removes the influence of *Z* when investigating the causal effect of *Y* on *X*. First, the optimal time delay of the causal influence from *Z* to *X* is determined by examining all possible time delays between variables via Eq. (13) and selecting the time delay that exhibits the highest causal strength [39], which can be expressed as:

$$\xi_{ZX} = \arg\max_{\xi \geq 0}\left(\rho_{Z \to X, \xi}\right) \quad (14)$$

, and the corresponding predicted time series $\hat{\mathbf{z}}_{t_{\min}:L-\xi_{ZX}}^{X \to Z, \xi_{ZX}}$ can be obtained via Eq. (12), denoted as $\hat{Z}_X$.

Then, the optimal time delay of the influence from $Y$ to $\hat{Z}_X$ is obtained similarly by:

$$\xi_{Y\hat{Z}_X} = \arg\max_{\xi \geq 0}\left(\rho_{Y \to \hat{Z}_X, \xi}\right) \quad (15)$$

, and the corresponding predicted time series $\hat{\mathbf{y}}_{t_{\min}:L-\xi_{ZX}-\xi_{Y\hat{Z}_X}}^{X \to Z \to Y; \xi_{ZX}, \xi_{Y\hat{Z}_X}}$ can be derived using Eq. (12). If time series $Y$ exhibits a high similarity with $\hat{\mathbf{y}}_{t_{\min}:L-\xi_{ZX}-\xi_{Y\hat{Z}_X}}^{X \to Z \to Y; \xi_{ZX}, \xi_{Y\hat{Z}_X}}$, it suggests a causal influence along $Y \to Z \to X$. Following this, Eq. (9) can be extended as:

$$\rho_{Y \to X|Z} = \text{ppcc}\left(\hat{\mathbf{y}}_{t_{\min}:L-\xi_{YX}-\xi_{Y\hat{Z}_X}}^{X \to Y, \xi_{YX}}, \mathbf{y}_{t_{\min}:L-\xi_{ZX}-\xi_{Y\hat{Z}_X}} \middle| \hat{\mathbf{y}}_{t_{\min}:L-\xi_{ZX}-\xi_{Y\hat{Z}_X}}^{X \to Z \to Y; \xi_{ZX}, \xi_{Y\hat{Z}_X}}\right) \quad (16)$$

where $\text{ppcc}(\mathbf{e}_1, \mathbf{e}_2 | \mathbf{e}_3)$ denotes the partial PCC calculation quantifying the correlation between two vectors $\mathbf{e}_1$ and $\mathbf{e}_2$ when considering the vector $\mathbf{e}_3$; $\xi_{YX}$ denotes the optimal time delay of the causal influence from $Y$ to $X$; and $\rho_{Y \to X|Z}$ denotes the direct causal strength from $Y$ to $X$ considering $Z$ calculated by PCM. If the value $\rho_{Y \to X|Z}$ exceeds the predefined threshold $c$, it means that $Y$ has a direct causal effect on $X$ considering $Z$.

2.4 Problem Formulation

Consider a KPI that needs to be predicted, denoted as $X$. The monitored sensor data includes $M$ auxiliary variables $Y_1, Y_2, \ldots Y_M$, each of which records historical measurements over $N$ time points. Specifically, for the KPI at time $l$, denoted as $x(l)$, it can be predicted by:

$$\hat{x}(l) = f\left(\begin{bmatrix} y_1(l-1) & y_1(l-2) & \ldots & y_1(l-N) \\ \ldots & \ldots & \ldots & \ldots \\ y_M(l-1) & y_M(l-2) & \ldots & y_M(l-N) \end{bmatrix}\right) \quad (17)$$

where $y_i(j)$ represents the measurement of the $i^{\text{th}}$ auxiliary variable at time point $j$; and $\hat{x}(j)$ represents the prediction of the KPI at time point $j$. The aim of soft sensor modeling is to determine the mapping relationship $f$ between the auxiliary variables and the KPI. This paper focuses on the problem of feature selection as illustrated in Fig. 2, aiming to select features from the $M \times N$ feature set, which involves measurements of different variables at various time points, to improve the performance and stability of the soft sensor model.

Notably, some existing studies used historical measurements of the KPI as input features during the operation of industrial processes [40, 41]. However, in this study, we consider the most challenging scenario, where the KPI is entirely unobservable during operations, and is only observable in the laboratory for soft sensor model establishment. This is also the most common situation in practice.

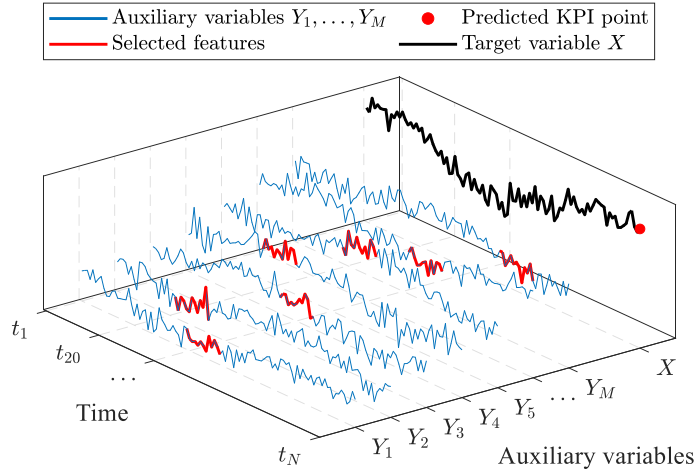

Fig. 2 Schematic diagram of the feature selection results.

## 3 Methodology

### 3.1 Motivation

This study aims to perform feature selection on auxiliary variable measurements recorded at multiple historical time points for soft sensor modeling. Quantifying the causal strength of each auxiliary variable at various time delays with respect to the current value of the KPI can provide valuable guidance for feature selection. However, as discussed in the introduction, existing causality-based feature selection methods investigate causal relationships in a same time dimension. Besides, variables in industrial processes are often interdependent, which contradicts the decorrelation assumption of traditional causal inference methods. To address these issues, we propose causal feature selection framework in this study based on time-delayed cross mapping.

For existing techniques, although TDCCM quantifies causal strengths across different time dimensions, it fails to distinguish direct causal relationships between variables. Consequently, TDCCM may introduce redundant features or ignore significant features, thereby impairing the stability of the soft sensor model. While PCM offers a way to quantify direct causality, it focuses only on the maximum causal strength across time delays, rather than the causal strength at specific lags, which limits its effectiveness for guiding feature selection in the time dimension. Therefore, in this section, we first propose TDPCM to infer direct causal relationships at different time delays within the state space reconstruction framework. Then, to achieve automatic causal feature selection, an objective selection strategy is proposed involving causal threshold optimization based on the model performance on the validation set. An outline of the proposed method is illustrated in Fig. 3.

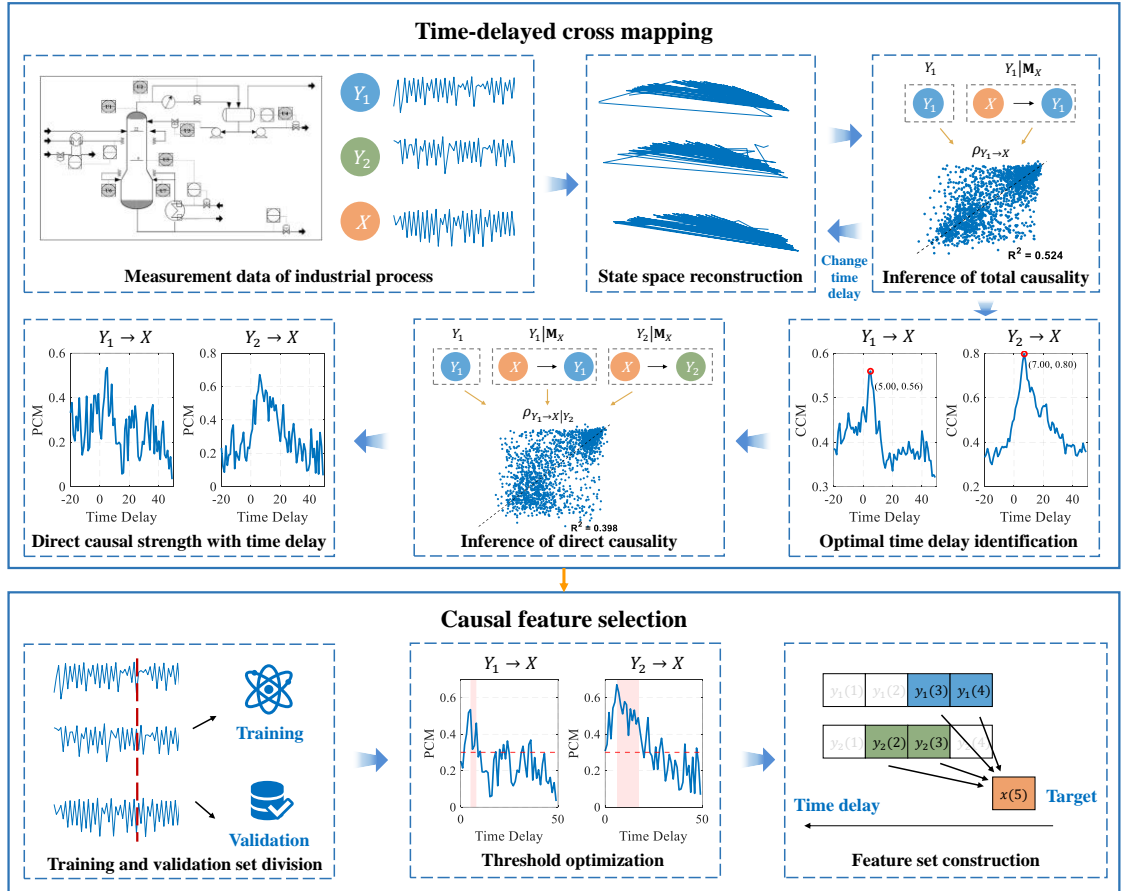

Fig. 3 Outline of the proposed method.

3.2 Time-Delayed Partial Cross Mapping

In this section, we present the TDPCM to quantify direct causal strength between interdependent time series. Without loss of generality, we consider a system with three time series, i.e., $X$, $Y$ and $Z$. Suppose that $Z$ exists in the causal path from $Y$ to $X$, indicating that the causal effect of $Y$ on $X$ is indirect, while that of $Z$ on $X$ is direct, as shown in Fig. 1 (b). At this point, if a unique causal influence delay $\xi_{YX}$ exists for the effect of $Y$ on $X$, then there must exist a corresponding delay $\xi_{YZ}$ on the intermediate path from $Y$ to $Z$ such that $\xi_{YZ} \leq \xi_{YX}$. When multiple delays from $Y$ to $X$ exist, there always exists at least one delay $\xi_{YX}$ for which a corresponding $\xi_{YZ}$ satisfying $\xi_{YZ} \leq \xi_{YX}$ exists. If the causal effect of $Y$ on $X$ can be covered by the causal effect of $Y$ on $Z$, it indicates that the causal effect of $Y$ on $X$ is indirect. Based on the above analysis, by introducing the predicted time series of $Y$ based on $\mathbf{M}_Z$ with the optimal time delay $\xi_{YZ}$, the influence of the confounding variable $Z$ can be eliminated.

Notably, Sugihara et al. [30] pointed out that when the unidirectional causal relationship between variables is excessively strong, it can lead to a phenomenon known as synchrony. This synchrony may produce a spurious bidirectional causal link, thereby interfering with the identification of the optimal time delay with the maximum causal strength. The TDCCM method can distinguish such spurious relationships [38]. Specifically, a negative time delay at the peak of causal strength indicates synchrony-induced false causality, whereas a delay of zero or greater suggests a valid causal relationship at the corresponding time lag. As a result, to mitigate the interference of synchrony, candidate time delays are restricted to the set of local maxima of the TDCCM curve. A local maximum at delay $\xi$ is defined as a

point satisfying:

$$\rho_{Y\to X,\xi} > \rho_{Y\to X,\xi-1}, \quad \rho_{Y\to X,\xi} > \rho_{Y\to X,\xi+1} \tag{18}$$

Let $\Xi_{YX}$ represent the set of time lags at which the causal influence calculated by TDCCM from $Y$ to $X$ reaches local maxima. Then, the optimal time delay that has the maximum causal strength from $Y$ to $X$ can be calculated by:

$$\xi_{YX} = \arg \max_{\xi \geq 0, \xi \in \Xi_{YX}} \left( \rho_{Y\to X,\xi} \right) \tag{19}$$

By limiting the optimal time delay to local maxima with values exceeding zero in Eq. (19), false causal delays introduced by synchrony at early time delays can be excluded. Similarly, the optimal time delay from $Y$ to $Z$ along the causal path $Y \to Z \to X$ can be determined by:

$$\xi_{YZ}^X = \arg \max_{0 \leq \xi \leq \xi_{YX}, \xi \in \Xi_{YZ}} \left( \rho_{Y\to X,\xi} \right) \tag{20}$$

where $\xi_{YZ}^X$ is the optimal time delay from $Y$ to $Z$ along the causal path $Y \to Z \to X$, which is less than $\xi_{YX}$.

Then, inspired by Eq. (16), the direct causal strength from $Y$ to $X$ at time delay $\xi_{YX}$ can be expressed as:

$$\rho_{Y\to X|Z,\xi_{YX}} = \mathrm{ppcc}\left( \hat{\mathbf{y}}_{t_{\min}:L-\xi_{YX}}^{X\to Y,\xi_{YX}}, \mathbf{y}_{t_{\min}:L-\xi_{YX}} \middle| \hat{\mathbf{y}}_{t_{\min}:L-\xi_{YZ}^X}^{Z\to Y,\xi_{YZ}^X} \right) \tag{21}$$

Equation (21) indicates that if the causal influence of $Y$ on $X$ at the optimal time delay $\xi_{YX}$ can be weakened by considering the causal influence of $Y$ on $Z$ at the optimal time delay $\xi_{YZ}^X$, it suggests that the causal link from $Y$ to $X$ is affected by $Z$. Based on Eq. (21), when considering a varying time delay $\gamma$ from $Y$ to $X$ considering $Z$, TDPCM can be described by:

$$\rho_{Y\to X|Z,\gamma} = \mathrm{ppcc}\left( \hat{\mathbf{y}}_{t_{\min}:L-\gamma}^{X\to Y,\gamma}, \mathbf{y}_{t_{\min}:L-\gamma} \middle| \hat{\mathbf{y}}_{t_{\min}:L-\gamma}^{Z\to Y,\xi_{YZ}^X-\xi_{YX}+\gamma} \right) \tag{22}$$

where $\rho_{Y\to X|Z,\gamma}$ denotes the direct causal strength from $Y$ to $X$ considering $Z$ at time delay $\gamma$ calculated by TDPCM. TDPCM is also influenced by the synchrony-induced false causality, therefore the direct causal influence delay from $Y$ to $X$ calculated by TDPCM is determined by:

$$\gamma_{YX} = \arg \max_{\gamma \geq 0, \gamma \in \Pi_{YX}} \left( \rho_{Y\to X,\gamma} \right) \tag{23}$$

where $\Pi_{YX}$ represents the set of time lags at which the causal influence calculated by TDPCM from $Y$ to $X$ reaches local maxima.

It should be noted that Eq. (22) only considers a single potential disturbing variable. If $N$ potential disturbing variables $\mathbf{Z} = \{Z_1, Z_2, \ldots, Z_Q\}$ exist, TDPCM can be extended by jointly considering all the predicted time series obtained by disturbing variables. First, we define a stacked vector at time delay $\gamma$ as:

$$\mathbf{U}_\gamma = \left[ \hat{\mathbf{y}}_{t_{\min}:L-\gamma}^{X\to Y,\gamma}, \mathbf{y}_{t_{\min}:L-\gamma}, \hat{\mathbf{y}}_{t_{\min}:L-\gamma}^{Z_1\to Y,\xi_{YZ_1}^X-\xi_{YX}+\gamma}, \ldots, \hat{\mathbf{y}}_{t_{\min}:L-\gamma}^{Z_N\to Y,\xi_{YZ_N}^X-\xi_{YX}+\gamma} \right] \tag{24}$$

Let $\mathbf{\Sigma}_\gamma = \mathrm{Cov}(\mathbf{U}_\gamma)$ be the full covariance matrix of $\mathbf{U}_\gamma$ and $\mathbf{\Omega}_\gamma = \mathbf{\Sigma}_\gamma^{-1}$ be its precision matrix. Then, TDPCM considering multiple disturbing variables can be calculated by:

$$\rho_{Y\to X|\mathbf{Z},\gamma} = \frac{(\mathbf{\Omega}_\gamma)_{12}}{\sqrt{(\mathbf{\Omega}_\gamma)_{11}(\mathbf{\Omega}_\gamma)_{22}}} \tag{25}$$

Compared to PCM given by Eq. (16), TDPCM given by Eqs. (22) and (24) has the following improvements:

- PCM eliminates the indirect causality by considering the predicted trajectory matrix of disturbing variables. For complex industrial processes with numerous variables, however, it necessitates examining all possible causal paths [39], significantly increasing computational efforts. For example, if four time series $W$, $X$, $Y$, $Z$ exist, when considering the causal impact of $W$ on $Z$, the following causal paths need to be considered for PCM: $W \rightarrow X \rightarrow Z$, $W \rightarrow Y \rightarrow Z$, $W \rightarrow X \rightarrow Y \rightarrow Z$ and $W \rightarrow Y \rightarrow X \rightarrow Z$. On the contrary, for TDPCM, the indirect causality is eliminated by considering the trajectory matrices of the disturbing variables directly, making it more suitable for applications in multivariable industrial processes. For the above example, the following causal paths need to be considered for TDPCM: $W \rightarrow X$ and $W \rightarrow Y$. A comparison of their computational complexity is given in Section 3.4.
- PCM emphasizes the maximum causal strength across all time delays rather than the causal strength at a specific one, limiting its ability to select features from various time points. In contrast, TDPCM quantifies the direct causal strength across all time delays, facilitating causal feature selection for soft sensor modeling. Besides, the time delay constraint in the propagation of causal influence among variables and the synchrony-induced false causality are both considered in TDPCM.

The procedures for calculating TDCCM and TDPCM are summarized in Algorithm 1.

**Remark 1**: The hyper-parameters for time-delayed cross mapping are the embedding dimension $E$ and the embedding time delay $\tau$. In this study, the embedding dimension is computationally determined using the false nearest neighbor (FNN) method [42]. Since the sampling interval of industrial time series is typically longer than its intrinsic dynamics, $\tau$ is set to 1. Sensitivity analysis is also conducted in Section 4.1.4 to verify the effectiveness of parameter settings.

**Remark 2**: Although TDPCM relies on the partial PCC which is a linear method, its combination with state space reconstruction is often adequate for general industrial processes. State-space reconstruction can transform nonlinear system dynamics into a higher-dimensional space, in which causal relationships can be manifested in an approximately linear form, especially for processes with smooth dynamics and relatively stable operating regimes. Nevertheless, when strong nonlinear confounding remains after reconstruction, the direct causal strength estimated by TDPCM may be biased, and its result should be interpreted with caution in such cases.

---

**Algorithm 1**: TDCCM and TDPCM calculation.

**Input**:
1. Time-series dataset containing the KPI $X$ and auxiliary variables $Y_1, Y_2, \ldots Y_M$
2. Maximum time delay $d$

**Output**:
1. The causal strength of all auxiliary variables to the KPI across the given time delay range as calculated by TDCCM and TDPCM

**Procedure**:
\# Compute TDCCM for all variable pairs
1. **for** each ordered pair $(i, k)$ with $i \neq k$ **do**:
2.     Compute $\rho_{Y_i \rightarrow Y_k, \xi}$ via Eq. (13), $\xi = 0, \ldots, d$
3. **end for**
4. **for** each auxiliary variable $Y_j$ **do**:
5.     Compute $\rho_{Y_j \rightarrow X, \xi}$ via Eq. (13), $\xi = 0, \ldots, d$

6.    **end for**
\# Determine optimal delays for $Y_j \to X$
7.    **for** each auxiliary variable $Y_j$ **do**:
8.        Determine the set of time lags $\boldsymbol{\Xi}_{Y_j X}$ at which the causal influence calculated by TDCCM from $Y_j$ to $X$ reaches local maxima
9.        Determine $\xi_{Y_j X}$ via Eq. (19)
10.  **end for**
\# Determine optimal delays $Y_j \to Y_i$ for disturbing variables
11.  **for** each auxiliary variable $Y_j$ **do**:
12.       Identify disturbing variable set $\mathbf{Y}_{\backslash j} = \{Y_i \mid i=1,\ldots,M, i \neq j\}$
13.       **for** each $Y_i \in \mathbf{Y}_{\backslash j}$ **do**:
14.           Determine the set of time lags $\boldsymbol{\Xi}_{Y_j Y_i}$ at which the causal influence calculated by TDCCM from $Y_j$ to $Y_i$ reaches local maxima
15.           Determine $\xi_{Y_j Y_i}$ via Eq. (20)
16.       **end for**
17.  **end for**
\# Compute TDPCM considering multiple disturbing variables for $Y_j \to X$
18.  **for** each auxiliary variable $Y_j$ **do**:
19.       Construct stacked vector $\mathbf{U}_\gamma$ via Eq. (24), $\gamma = 0, \ldots, d$
20.       Compute covariance and precision matrix
21.       Compute $\rho_{Y \to X \mid \mathbf{Y}_{\backslash j}, \gamma}$, $\gamma = 0, \ldots, d$
22.  **end for**

3.3  Causal Feature Selection

After developing the time-delayed cross mapping techniques, the next key issue is determining causal features for industrial soft sensor modeling based on their results. Typically, an empirical threshold is employed by analysts and features whose strength exceeds the threshold are added to the inputs. However, this empirical approach depends heavily on expert experience and is difficult to justify rigorously. If the threshold is set too low, redundant features may be introduced; if set too high, critical variables may be excluded. To overcome these limitations, we propose a data-driven method to determine the selection threshold objectively. The available data are divided into a training set and a validation set, where the training set is used to construct the soft sensor model and the validation set is used to optimize the selection threshold. The proposed method is built upon the following assumptions:

(1) **Validation performance reflects feature suitability**

A feature subset is considered more appropriate if it leads to better predictive performance on the validation set. This assumption is widely adopted in data-driven modeling and provides an objective criterion for comparing candidate feature sets. In this study, root mean square error (RMSE) is used to evaluate model performance during parameter optimization. RMSE is straightforward to interpret, computationally simple, and sensitive to outliers, making it well-suited for assessing soft sensor models. If feasible, other performance metrics can also be considered as alternatives.

(2) **Stronger causal strength implies greater relevance**

A larger causal strength indicates a stronger causal effect of an auxiliary variable on the KPI at a particular time delay. Therefore, features associated with higher causal strength should be prioritized in the selection process.

(3) **The soft sensor model is deterministic without inherent uncertainty**

Given a fixed input feature set, soft sensor provides a deterministic prediction of the KPI. This assumption ensures that model performance reflects the suitability of the chosen features rather than stochastic effects arising from model uncertainty. Consequently, the validation performance can be reliably used to optimize the causal strength threshold. In this study, due to the generally limited data available for soft sensor modeling and the repeated training required to determine optimal setting values, PLS is employed because it trains quickly and has demonstrated reliable performance in soft sensor modeling [4, 5]. If feasible, other models can also be considered as alternatives. For other regression models that involve inherent uncertainty, such as deep learning models, how to combine time-delayed cross mapping methods with them into the design of soft sensors will be a key focus of future research. Model adaptability analysis is also conducted in Section 4.1.4 using Gaussian-kernel support vector regression.

(4) **Historical measurements preceding the causal delay do not contribute to the KPI**

Historical measurements with time lags shorter than the causal influence delay identified by time-delayed cross mapping methods are assumed not to affect the KPI. For example, if variable $Y$ affects variable $X$ at a lag of five time steps, then measurements of $Y$ at time delays one to four relative to time point $l$ do not contribute to predicting $x(l)$. This assumption prevents the inclusion of non-informative or weakly related measurements during feature selection. Ablation study on this constraint is also conducted in Section 4.1.4.

Based on the above assumptions, the feature set is constructed based on causal threshold optimization, which consists of continuous historical measurements of each auxiliary variable starting from the identified causal time delay. The procedures for feature selection are presented as follows:

*Step 1: Initialization*. Split the given dataset into training and validation sets. Specify the maximum time delay $d$ and the value space for the index threshold as $\mathbf{C} = \{c_1, c_2, \ldots, c_D\}$.

*Step 2: Feature set construction*. For each index threshold $c_i$, the continuous historical measurements of auxiliary variables with TDPCM values greater than $c_i$ and time delays larger than causal influence delay are selected as model inputs for soft sensor modeling. A soft sensor model is then constructed using the training set, and its performance is evaluated on the validation set to obtain the performance corresponding to $c_i$. After evaluating the model performance across all possible index thresholds, the threshold that yields the best performance is selected as the optimal index threshold, denoted $c_{best}$. And, the features selected based on $c_{best}$ constitute the feature set.

The above feature selection procedures are summarized in Algorithm 2.

**Remark 3**: A predefined threshold candidate set $\mathbf{C}$ is required before training. To construct this, we propose a simple and objective approach: select $D - 1$ equally spaced values between the lowest and highest TDPCM values obtained across all auxiliary variables, and additionally include the maximum TDPCM value for each individual auxiliary variable. This approach integrates a global perspective while simultaneously capturing local characteristics. As a result, the constructed threshold space is both comprehensive and representative, facilitating effective feature selection.

---

**Algorithm 2**: Feature selection based on the results of time-delayed cross mapping.
**Input**:
1. Time-series dataset containing the KPI $X$ and auxiliary variables $Y_1, Y_2, \ldots Y_M$

2. Threshold candidate set $\mathbf{C} = \{c_1, c_2, \ldots, c_D\}$
3. Maximum time delay $d$
4. TDCCM and TDPCM values from all auxiliary variables to the KPI across given time delays

**Output**:
1. Selected features for soft sensor modeling

**Procedure**:
1. **for** each index threshold $c_i \in C$ **do**:
2.     Initialize selected feature set $\mathcal{F}(c_i) \leftarrow \emptyset$
3.     **for** each auxiliary variable $Y_j$ **do**:
4.       Determine the causal influence delay $\delta_j$ via Eq. (19) for TDCCM and Eq. (23) for TDPCM
5.       Set $\gamma \leftarrow \delta_j$
6.       **while** $\gamma \leq d$ and $\rho_{Y \to X, \gamma} \geq c_i$ for TDCCM, or $\gamma \leq d$ and $\rho_{Y \to X | \mathbf{Y}_{\backslash j}, \gamma} \geq c_i$ for TDPCM **do**
7.         $\mathcal{F}(c_i) \leftarrow \mathcal{F}(c_i) \cup \{y_j(t - \gamma)\}$
8.         $\gamma \leftarrow \gamma + 1$
9.       **end while**
10.     **end for**
11.     Train a soft sensor model using $\mathcal{F}(c_i)$ on the training set
12.     Evaluate its performance on the validation set and record the score $J(c_i)$
13. **end for**
14. Select the optimal threshold $c_{best} = \arg\min_{c_i \in \mathbf{C}} J(c_i)$
15. Return the final selected feature set $\mathcal{F}(c_{best})$

### 3.4 Complexity Analysis

First, the computational complexity of TDCCM is analyzed. For a single TDCCM computation at a given time delay, $E + 1$ nearest neighbors are identified in the reconstructed space for each embedding vector, and the effective number of embedding vectors approaches $L^* = L - (E-1)\tau$. Therefore, considering $d + 1$ different time delays, the complexity per variable pair scales as $O(dEL^*)$. The construction of trajectory matrices and the calculation of PCC introduce additional computational costs of $O(L^*)$ per time delay, which are negligible compared to the nearest neighbor search. Since there are $M$ auxiliary variables in total, the overall computational complexity of TDCCM is $O(M^2 dEL^*)$. In practice, since $E$ is usually much smaller than $L^*$, the complexity can be approximately regarded as $O(M^2 dL^*)$.

After performing TDCCM analysis, the optimal delays between variables are selected. Then, TDPCM employs the PPCC to calculate the direct causal strength between auxiliary variables and the KPI. Since the costs of delay selection and PPCC computation are negligible, the overall computational complexity of the TDPCM is dominated by that of TDCCM.

Then, we analyze the computational complexity of the feature selection strategy. Let the computational complexity of a single training and validation of the soft sensor model with $m$ input features be denoted as $T_{model}(m)$. In Step 2, the soft sensor model is trained and validated for each candidate threshold $c_i$. A total of $k$ model training and validation operations are conducted in this step. Therefore, the computational complexity of the proposed feature selection strategy is $O(kT_{model}(\bar{m}))$, where $\bar{m}$ represents the average number of selected features after screening.

**Remark 4**: If the original calculation method of PCM is used, where the predicted trajectory matrix of disturbing variables must be constructed, additional steps are required, including nearest-neighbor searches and exploration of all possible causal paths. For a system with $M$ variables, the maximum depth of a causal path can reach $M - 2$. At this point, after performing TDCCM analysis, the computational complexity of PCM becomes $O(M^{M-2}dL^*)$ when considering all time delays, which is much higher than the computational complexity of the proposed method.

## 4 Experiments

In this section, two real-world industrial cases are introduced to implement the proposed method. To show the superiority of the proposed feature selection method based on time-delayed cross mapping, the following methods are employed for comparison:

- Feature selection approaches that ignore the impact of time delay, including correlation-based method PCC [10] and causality-based method CMI [22]. For both methods, it is necessary to determine appropriate threshold values for their respective metrics, as well as the time range to be considered. These parameters are optimized based on the model performance on the validation set.

- State-of-the-art model training-based feature selection method RF [14]. For this method, we rank the feature importance and select features based on the model performance on the validation set.

- Feature selection approaches based on baseline time-series causal inference methods with the time-delay framework, including TDGC and TDTE. For these methods, the proposed feature selection strategy is employed.

- Original soft sensor model without any feature selection methods, which is also known as the sliding window technique for time series prediction [43].

The implementation details of all the feature selection methods are provided in Appendix A. In order to compare the performance of soft sensor models, three widely used metrics are employed: coefficient of determination ($R^2$), RMSE and mean absolute error (MAE). For all the feature selection methods, PLS is used for soft sensor modeling.

### 4.1 Debutanizer Column

#### 4.1.1 Dataset Description

The debutanizer column plays an important role in petroleum refining, particularly in naphtha cracking and desulfurization. The flowchart of the debutanizer column is illustrated in Fig. 4, with detailed descriptions of monitoring variables provided in Table 1. Specifically, U1 represents the temperature at the top of the column, directly influencing vapor composition and condensation behavior. U2 indicates the pressure at the overhead of the column, affecting both the purity of separated products and overall separation efficiency. U3 controls the liquid reflux returned to the column, significantly impacting product purity and operational stability. U4 describes the rate at which the top product is transferred to subsequent refining stages, reflecting process throughput. U5 provides a mid-column temperature reading, which can be regarded as a sensitive indicator of internal separation efficiency and composition gradients. U6 and U7 measure temperatures in the bottom section of the column. These are essential for managing the vaporization conditions and ensuring the minimized butane content.

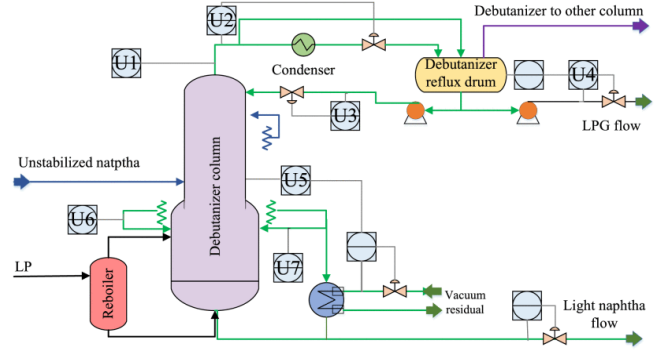

Fig. 4 Flowchart of the debutanizer column.

Table 1 Monitoring variables on the debutanizer column case.

| Monitoring variables | Variable description | Unit |
|---|---|---|
| U1 | Top Temperature | °C |
| U2 | Top pressure | kg/cm² |
| U3 | Reflux flow | m³/h |
| U4 | Flow to next process | m³/h |
| U5 | 6th tray temperature | °C |
| U6 | Bottom temperature A | °C |
| U7 | Bottom temperature B | °C |

Minimizing the butane content in the bottom of the distillation column is crucial for enhancing product quality. However, in practical applications, the gas chromatography method commonly used to measure butane content suffers from significant time delays, which impedes real-time system control and thus affects product performance. Therefore, it is necessary to establish a soft sensor model to estimate butane content online. To achieve this, the above seven easily measurable auxiliary variables physically related to the butane content are recorded, denoted as U1-U7, and the butane content is denoted as U8, which is the KPI to be predicted. Further details on debutanizer column can be found in [44].

In this study, there are a total of 2194 samples gathered from the process, sourced from [44]. The first 1596 samples are used for causal inference and model establishment, and the remaining 598 samples are reserved for model testing. Before analysis, all variables are scaled using min-max normalization.

4.1.2 Causal Inference

As stated in Remark 1, the hyper-parameters for causal inference based on time-delayed cross mapping are the embedding dimension $E$ and the embedding time delay $\tau$, where $E$ is determined according to the FNN method, and $\tau$ is set to a default value of 1. Fig. 5 shows the variation of the FNN with the embedding dimension for all the variables in the debutanizer column case. An embedding dimension of $E = 4$ is selected, as it is the first dimension at which all FNN values drop below 5%.

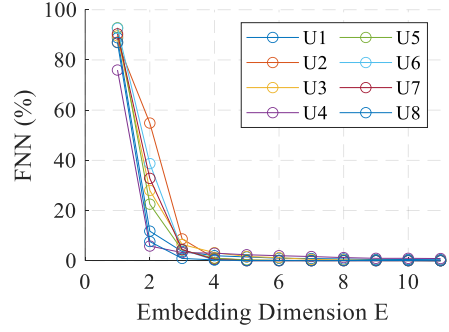

Fig. 5 Variation of FNN with increasing embedding dimension for different variables on the debutanizer column case.

Then, according to Algorithm 1, the causal inference results of TDCCM and TDPCM can be obtained. A negative lag search window of 50 is employed to identify potential synchrony-induced false causality. Fig. 6 presents the causal strength with time delay of each auxiliary variable on the KPI as calculated by TDCCM and TDPCM. It indicates that U1, U3 and U5 have strong causal impacts on the KPI, whereas U2, U4, U6 and U7 show relatively weak causal relationships. This result can be explained by process knowledge: U1 reflects the extent of butane removal; U3 determines the efficiency of mass transfer, which governs the separation process; and U5 provides information about the composition variation. They all have strong causal impacts on the KPI. In contrast, U6 and U7 are influenced by the KPI rather than affecting it, resulting in weak causal impacts. Although U2 is theoretically expected to have a strong causal impact on the KPI, its influence is not detected. This is likely because U2 is tightly constrained in the process, exhibiting very limited variability. As listed in Table 2, the coefficient of variation of U2 is only 0.043, far below those of other auxiliary variables. Compared with TDCCM, the causal strength of U3 and U5 calculated by TDPCM remains unchanged, indicating that their impacts on the KPI are predominantly direct. The above analysis demonstrates that the causal inference results show satisfactory consistency with the process knowledge.

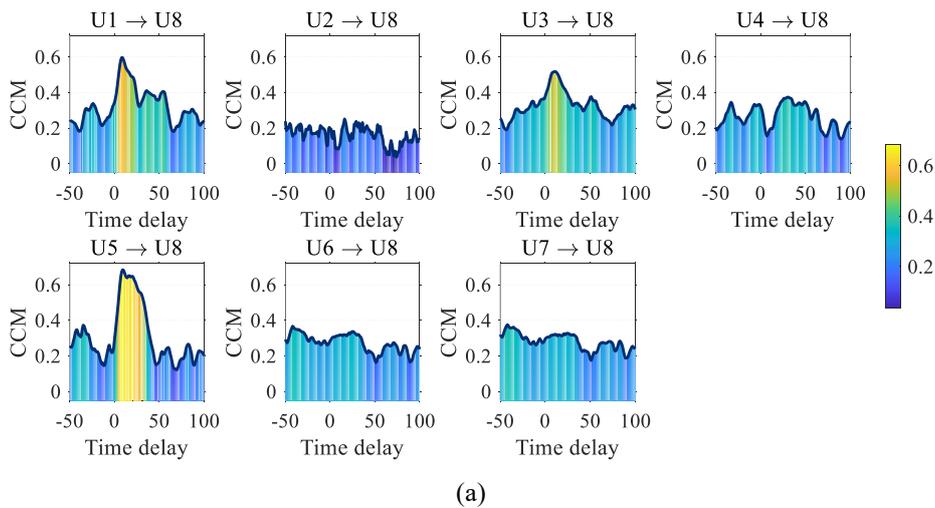

(a)

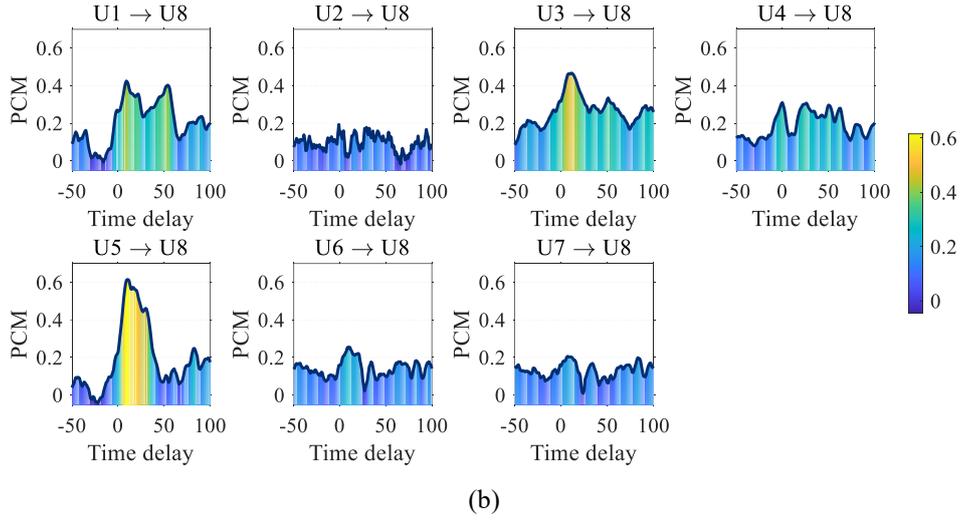

(b)

Fig. 6 Causal strength of each auxiliary variable on the KPI on the debutanizer column case: (a) TDCCM; (b) TDPCM.

Table 2 Coefficient of variation of auxiliary variables on the debutanizer column case.

| Monitoring variables | Coefficients of variation |
| --- | --- |
| U1 | 0.3552 |
| U2 | 0.0430 |
| U3 | 0.3652 |
| U4 | 0.2417 |
| U5 | 0.1498 |
| U6 | 0.2266 |
| U7 | 0.2549 |

Fig. 7 illustrates how PCM values between auxiliary variables and the KPI at the causal influence delay vary with increasing sample size. From Fig. 7, all auxiliary variables tend to converge toward stable values, showing the robustness of the causal inference results.

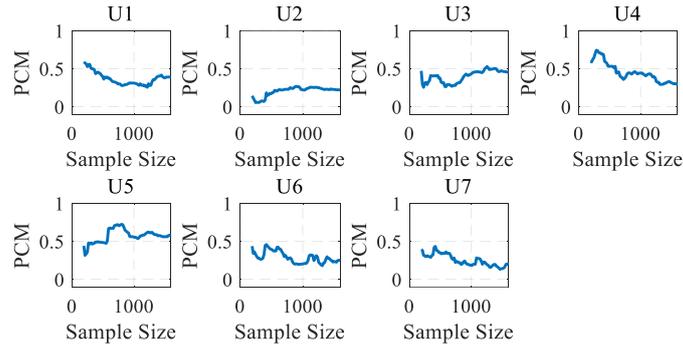

Fig. 7 Variation of PCM values between auxiliary variables and the KPI at the causal influence delay with increasing sample size.

4.1.3 Soft Sensor Modeling Results and Comparisons

Time-delayed causality analysis introduces a temporal shift between the auxiliary variables and the KPI. With a maximum time delay $d$ of 100, the first 100 samples of KPI cannot be labeled because the corresponding historical measurements of auxiliary variables are unavailable. As a result, 1496 effective labeled samples are available for model development. When developing soft sensor models, it is essential to confirm that the model trained on the training set maintains satisfactory performance on unseen future data. However, the training and validation sets are always divided randomly in practice, which can influence the model performance. If the model results are significantly affected by the dataset division, it suggests weak model stability and introduces substantial risks in practical applications. To systematically evaluate the stability of different methods, we vary the sizes of the training set from 1000 to 1400 samples in steps of 1, while keeping the total number of training and validation samples fixed at 1496. Notably, to avoid information leakage in the prediction task, the time series samples are not shuffled. Instead, a chronological cutoff point is selected on the time axis, and all observations before the cutoff form the training set while the remaining observations form the validation set. In this way, the temporal order of the data is strictly preserved, and models are always trained on past data and validated on future data. Taking a training set sample size of 1350 as an example, the data division strategy is shown in Fig. 8. The test set corresponds to the last segment of the time series and is used only for final performance evaluation, after causal feature selection, threshold determination and hyper-parameter tuning have been completed.

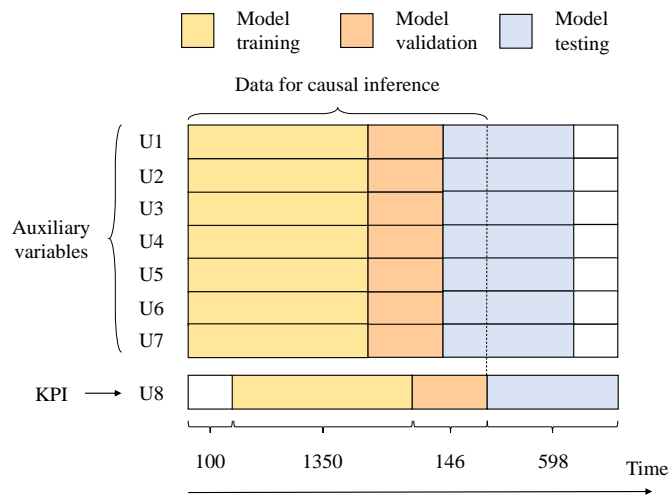

Fig. 8  Schematic diagram of the data division strategy.

For model establishment, PLS is used for soft sensor modeling combined with a specific feature selection method, with n_components = 3. The causal features are selected according to Algorithm 2. Taking a training set sample size of 1350 as an example, Fig. 6 highlights the selected features with the optimized index threshold by TDCCM and TDPCM. It can be seen that features from U1, U3, U4 and U5 are selected across a long time range, while features from other auxiliary variables are selected in a limited time range. Notably, some features from U6 and U7 are included, possibly due to their relevance to the historical state of the KPI.

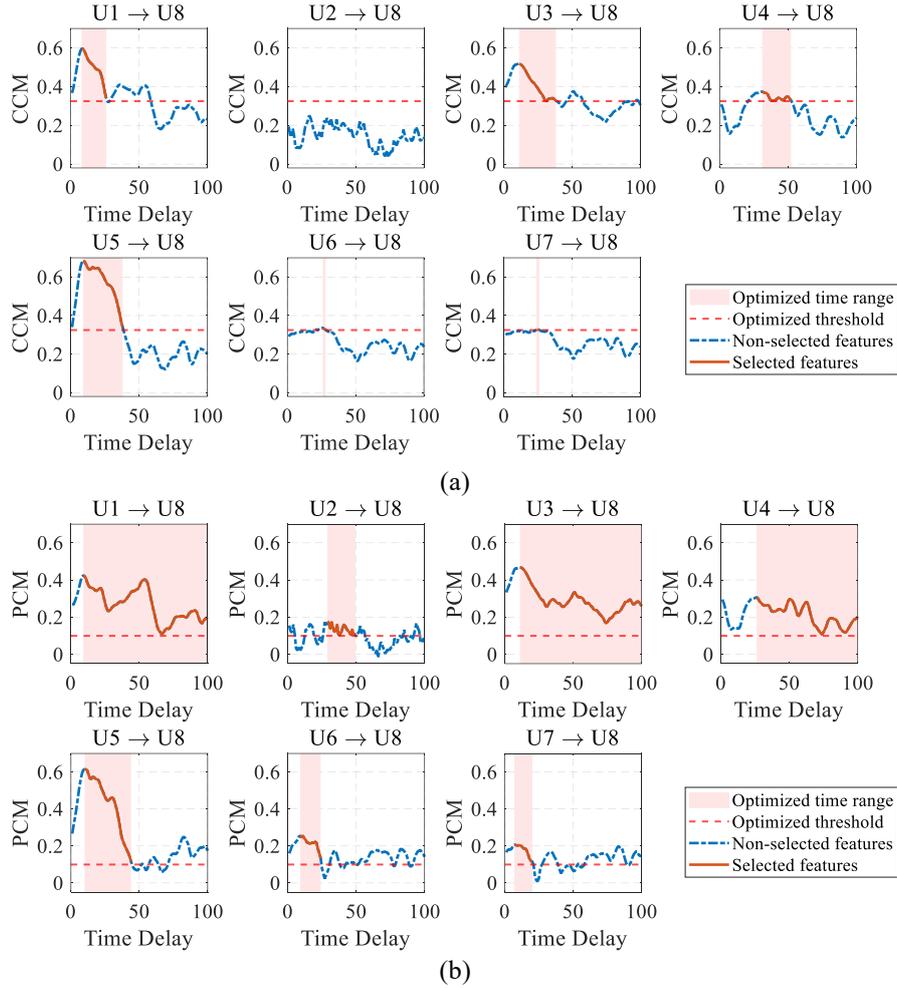

Fig. 9 Selected features of each auxiliary variable with optimized threshold on the debutanizer column case: (a) TDCCM; (b) TDPCM.

The performance metrics on the test set with increasing sample size of the training set using different feature selection methods are shown in Fig. 10. It can be seen that the performance on the test set does not monotonically increase with larger training set sizes but instead fluctuates for several feature selection methods. This behavior is primarily due to the presence of irrelevant or weakly relevant features. When feature selection fails to effectively remove disturbing or weakly relevant lagged measurements, the resulting models become sensitive to redundant or interfering information [45]. In such cases, different train-validation splits may include different subsets of these features, which leads to oscillatory performance curves. Overall, models based on RF and TDPCM exhibit greater stability compared to other feature selection methods, stabilizing around a sample size of 1300. By comparison, the performance of other feature selection methods is affected by the dataset division, resulting in relatively poor model stability.

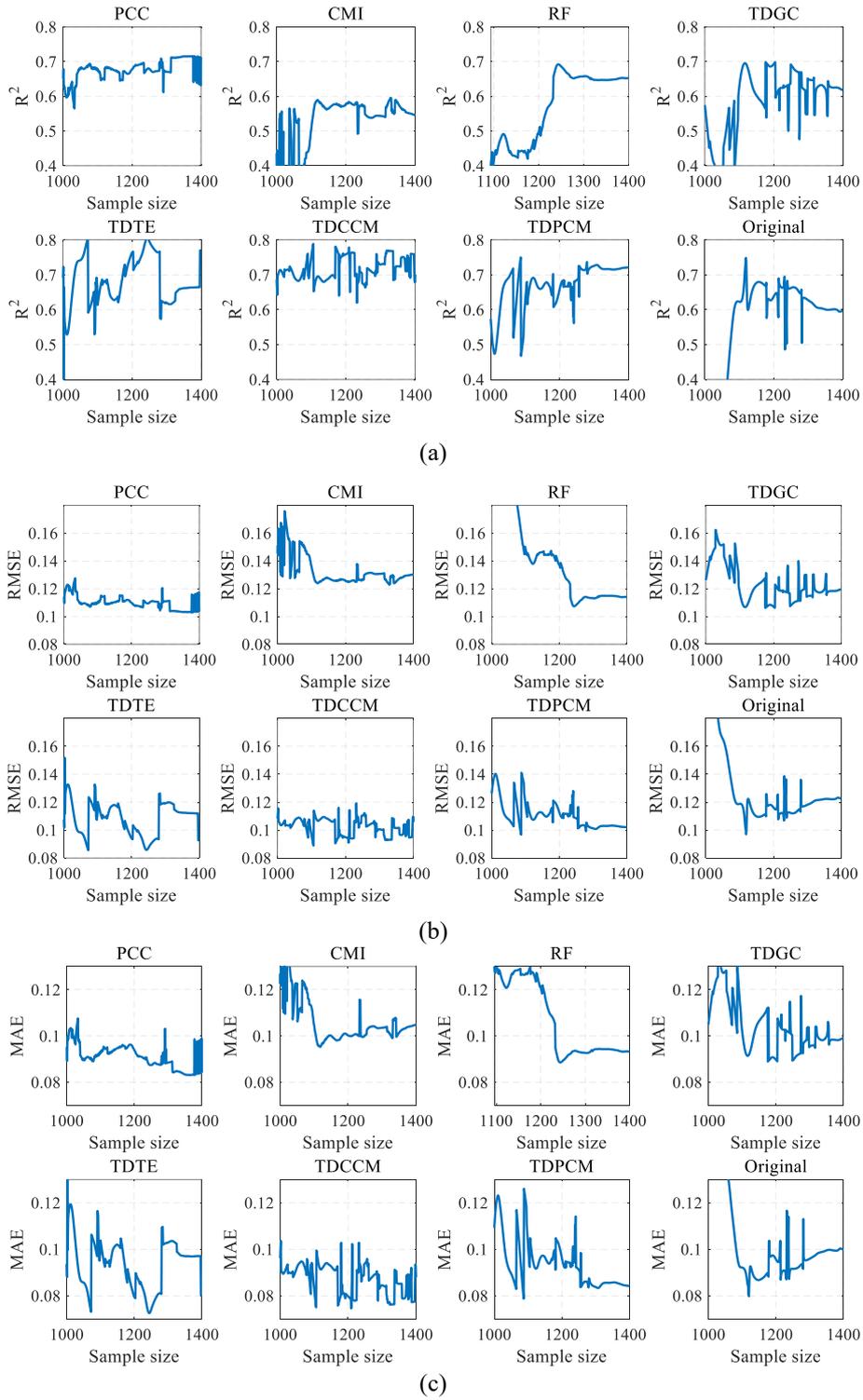

Fig. 10 Variation of performance metrics with increasing sample size of the training set using different feature selection methods on the debutanizer column case: (a) $R^2$ (b) RMSE (c) MAE.

Table 3 presents the average performance, standard deviation, the worst performance and the best performance on the test set using different feature selection methods when the sample size of the training set exceeds 1300. For model average performance, the performance of CMI is worse than that of other causal feature selection methods and the original model, demonstrating the critical role of incorporating time delay in identifying causal features. PCC demonstrates strong model accuracy, which suggests that correlation-based approaches may sometimes be suitable for feature selection. Compared to TDGC, which focuses on linear relationships, TDTE performs better by capturing nonlinear relationships. However, the performance of TDTE is still inferior to that of TDCCM due to its dependence on the decorrelation assumption, which shows the superiority of the state space reconstruction-based causal inference techniques in industrial processes. Among all methods, TDCCM achieves the best average performance, followed by TDPCM. Compared with the existing best method PCC, TDCCM yields a 5.33% increase in $R^2$, along with 6.53% and 4.66% reductions in RMSE and MAE, respectively. For model uncertainty, the standard deviation of RF is the best, followed by TDPCM. As the method with minimal risk, TDPCM achieves a 5.43% increase in $R^2$ and reductions of 5.93% in RMSE and 8.24% in MAE compared to the second-best method TDCCM in the worst scenario. Overall, the feature selection method based on time-delayed cross mapping techniques achieve superior average performance and stability compared to existing feature selection methods.

Table 3  Evaluation results of different feature selection methods on the debutanizer column case.

| Methods | $R^2$ | RMSE | MAE |
|---|---|---|---|
| PCC [10] | 0.7012±0.0229 (0.6317, 0.7156) | 0.1056±0.0039 (0.1031, 0.1173) | 0.0859±0.0042 (0.0831, 0.0988) |
| CMI [22] | 0.5613±0.0159 (0.5399, 0.5951) | 0.1280±0.0023 (0.1230, 0.1311) | 0.1032±0.0020 (0.0990, 0.1078) |
| RF [14] | 0.6501±**0.0034** (0.6466, 0.6579) | 0.1144±**0.0006** (0.1131, 0.1149) | 0.0937±**0.0006** (0.0926, 0.0944) |
| TDGC | 0.6243±0.0217 (0.5428, 0.6540) | 0.1185±0.0033 (0.1137, 0.1307) | 0.0985±0.0021 (0.0948, 0.1061) |
| TDTE | 0.6547±0.0302 (0.6158, **0.7704**) | 0.1135±0.0052 (**0.0929**, 0.1198) | 0.0981±0.0045 (<u>0.0803</u>, 0.1035) |
| TDCCM | **0.7386**±0.0253 (<u>0.6783</u>, <u>0.7694</u>) | **0.0987**±0.0047 (<u>0.0929</u>, <u>0.1097</u>) | **0.0819**±0.0049 (**0.0761**, <u>0.0934</u>) |
| TDPCM | <u>0.7191</u>±0.0039 (**0.7151**, 0.7284) | <u>0.1025</u>±0.0007 (0.1008, **0.1032**) | <u>0.0849</u>±0.0007 (0.0833, **0.0857**) |
| Original | 0.6048±0.0084 (0.5943, 0.6269) | 0.1215±0.0013 (0.1181, 0.1232) | 0.0988±0.0012 (0.0959, 0.1003) |

*Note*: The best results are highlighted in bold and the second-best results are underlined. The maximum and minimum values are reported in parentheses, respectively.

To statistically assess the performance differences between the proposed feature selection methods and the baseline approaches, the Wilcoxon signed-rank test is employed [46] using paired results obtained under identical training-validation configurations. The analysis focuses on the stable regime where the training sample size exceeds 1300, yielding 101 paired samples for each comparison. No multiple-comparison correction is applied, as the Wilcoxon tests are used as a supplementary analysis to support the observed performance differences.

In addition to statistical significance, the median difference (Median Δ) is reported to quantify the practical magnitude of the performance differences. Median Δ denotes the median performance improvement of the proposed method relative to the baseline across paired training-validation configurations, with positive values indicating better performance.

The results for TDCCM and TDPCM are reported in Tables 4 and 5. Across all performance metrics, existing feature selection methods differ significantly from TDCCM and TDPCM, highlighting the consistent superiority of our proposed approach.

Table 4  Wilcoxon signed-rank test and median difference results of TDCCM on the debutanizer column case.

| Metric | Comparison | R⁺ | R⁻ | p-value | Sig. (p<0.001) | Median Δ |
|---|---|---|---|---|---|---|
| $R^2$ | TDCCM vs PCC | 4911 | 240 | 1.29E-15 | + | 0.0405 |
| $R^2$ | TDCCM vs CMI | 5151 | 0 | 1.35E-18 | + | 0.1867 |
| $R^2$ | TDCCM vs RF | 5151 | 0 | 1.35E-18 | + | 0.0942 |
| $R^2$ | TDCCM vs TDGC | 5151 | 0 | 1.35E-18 | + | 0.1177 |
| $R^2$ | TDCCM vs TDTE | 5011 | 140 | 7.99E-17 | + | 0.0881 |
| $R^2$ | TDCCM vs TDPCM | 4468 | 683 | 7.3E-11 | + | 0.0258 |
| $R^2$ | TDCCM vs Original | 5151 | 0 | 1.35E-18 | + | 0.1372 |
| RMSE | TDCCM vs PCC | 238 | 4913 | 1.22E-15 | + | 0.0076 |
| RMSE | TDCCM vs CMI | 0 | 5151 | 1.35E-18 | + | 0.0306 |
| RMSE | TDCCM vs RF | 0 | 5151 | 1.35E-18 | + | 0.0166 |
| RMSE | TDCCM vs TDGC | 0 | 5151 | 1.35E-18 | + | 0.0203 |
| RMSE | TDCCM vs TDTE | 151 | 5000 | 1.09E-16 | + | 0.0158 |
| RMSE | TDCCM vs TDPCM | 665 | 4486 | 4.89E-11 | + | 0.0049 |
| RMSE | TDCCM vs Original | 0 | 5151 | 1.35E-18 | + | 0.0236 |
| MAE | TDCCM vs PCC | 877 | 4274 | 4.4E-09 | + | 0.0043 |
| MAE | TDCCM vs CMI | 0 | 5151 | 1.35E-18 | + | 0.0228 |
| MAE | TDCCM vs RF | 1 | 5150 | 1.39E-18 | + | 0.0131 |
| MAE | TDCCM vs TDGC | 0 | 5151 | 1.35E-18 | + | 0.0175 |
| MAE | TDCCM vs TDTE | 77 | 5074 | 1.31E-17 | + | 0.0182 |
| MAE | TDCCM vs TDPCM | 1128 | 4023 | 4.75E-07 | + | 0.0040 |
| MAE | TDCCM vs Original | 0 | 5151 | 1.35E-18 | + | 0.0182 |

Table 5  Wilcoxon signed-rank test and median difference results of TDPCM on the debutanizer column case.

| Metric | Comparison | R⁺ | R⁻ | p-value | Sig. (p<0.001) | Median Δ |
|---|---|---|---|---|---|---|
| $R^2$ | TDPCM vs PCC | 5150 | 1 | 1.39E-18 | + | 0.0047 |
| $R^2$ | TDPCM vs CMI | 5151 | 0 | 1.35E-18 | + | 0.1601 |
| $R^2$ | TDPCM vs RF | 5151 | 0 | 1.35E-18 | + | 0.0689 |
| $R^2$ | TDPCM vs TDGC | 5151 | 0 | 1.35E-18 | + | 0.0911 |
| $R^2$ | TDPCM vs TDTE | 5141 | 10 | 1.82E-18 | + | 0.0561 |
| $R^2$ | TDPCM vs Original | 5151 | 0 | 1.35E-18 | + | 0.1151 |
| RMSE | TDPCM vs PCC | 1 | 5150 | 1.39E-18 | + | 0.0009 |
| RMSE | TDPCM vs CMI | 0 | 5151 | 1.35E-18 | + | 0.0258 |
| RMSE | TDPCM vs RF | 0 | 5151 | 1.35E-18 | + | 0.0119 |
| RMSE | TDPCM vs TDGC | 0 | 5151 | 1.35E-18 | + | 0.0154 |
| RMSE | TDPCM vs TDTE | 97 | 5054 | 2.34E-17 | + | 0.0098 |
| RMSE | TDPCM vs Original | 0 | 5151 | 1.35E-18 | + | 0.0191 |
| MAE | TDPCM vs PCC | 2912 | 2239 | 0.8732 | / | -0.0010 |
| MAE | TDPCM vs CMI | 0 | 5151 | 1.35E-18 | + | 0.0189 |
| MAE | TDPCM vs RF | 0 | 5151 | 1.35E-18 | + | 0.0087 |
| MAE | TDPCM vs TDGC | 0 | 5151 | 1.35E-18 | + | 0.0133 |
| MAE | TDPCM vs TDTE | 10 | 5141 | 1.82E-18 | + | 0.0123 |
| MAE | TDPCM vs Original | 0 | 5151 | 1.35E-18 | + | 0.0138 |

Fig. 11 shows the comparison between actual values and predicted values of soft sensor models constructed with different feature selection methods when the training set sample size is 1350. The results indicate that the approaches based on time-delayed cross mapping achieve superior accuracy in overall prediction.

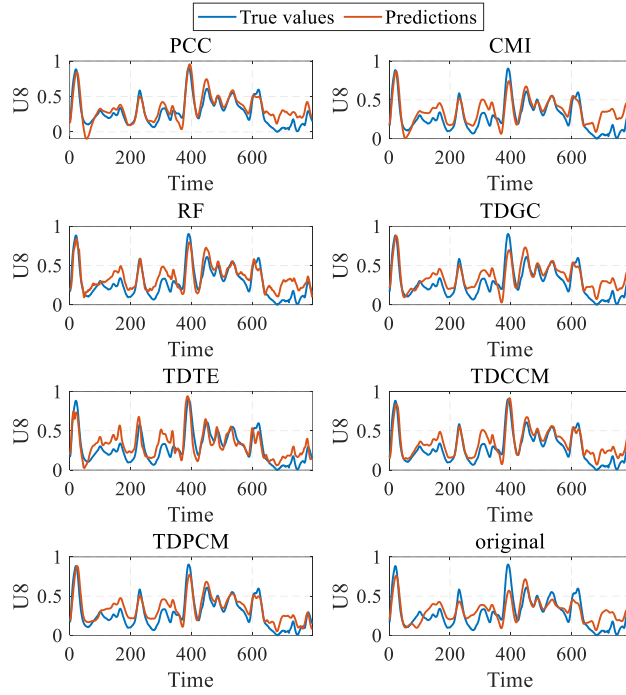

Fig. 11 Comparison between predicted and actual values for each method on the debutanizer column case.

Table 6 presents the detailed time-delayed features selected by each method when the training set sample size is 1350. For example, since TDPCM selects 7~42 for U1, the input features for predicting the KPI at time point $l$ include $u_1(l-7)$ to $u_1(l-42)$. It can be observed that TDGC-based method has the largest number of features, followed by TDPCM, while TDCCM has the lowest number of features. This suggests that the model performance is not determined by the mere number of features, and our approach improves model performance and stability by selecting features that involve causal information.

Table 6 Selected features of different methods on the debutanizer column case.

| Methods | U1 | U2 | U3 | U4 |
|---|---|---|---|---|
| PCC | / | 1~50 | 1~50 | / |
| CMI | 1~49 | / | 1~49 | 1~49 |
| RF | 1, 3, 5, 9, 11, 13, 15~16, 19~20, 22~24, 30~31, 39, 66, 72, 77, 79, 81~82, 88, 91~93, 97, 100 | / | 2~18, 22, 28~30, 62, 78~79, 96 | 1, 5~6, 9, 11~14, 21, 24~27, 29~38, 50, 52, 76~77, 82~83, 89~100 |
| TDGC | 7~62 | 13~100 | 5~42 | 12~60 |
| TDTE | 13~69 | / | / | 51~100 |
| TDCCM | 8~26 | / | 11~38 | 31~52 |
| TDPCM | 6~100 | 27~58 | 10~100 | 23~100 |
| Original | 1~100 | 1~100 | 1~100 | 1~100 |
| Methods | U5 | U6 | U7 | Total number |
| PCC | 1~50 | 1~50 | 1~50 | 250 |
| CMI | 1~49 | 1~49 | 1~49 | 294 |
| RF | 1~2, 11~16, 18~20, 22~24, 30, 43~44, | 60 | 52, 64, 89 | 125 |
| | 59, 65, 68, 70~71, 74, 80, 91, 95, 100 | | | |
| TDGC | 7~100 | 7~23 | 7~100 | 436 |
| TDTE | 6~14 | / | / | 116 |
| TDCCM | 9~38 | 26~28 | 24~26 | 105 |
| TDPCM | 7~42 | 9~24 | 10~19 | 358 |
| Original | 1~100 | 1~100 | 1~100 | 700 |

4.1.4 Discussions

*1) Sensitivity Analysis on Hyper-Parameters*

Tables 7 and 8 present the average performance and standard deviation on the test set using TDCCM and TDPCM under different parameter settings, respectively. The results indicate that TDCCM demonstrates low sensitivity to the $E$ and moderate sensitivity to $\tau$, whereas TDPCM is moderately sensitive to $E$ but highly sensitive to $\tau$. A notable drop in model accuracy occurs when $\tau$ surpasses 1, likely because industrial process data often have large sampling intervals. In such cases, an increased $\tau$ may hinder effective direct causality inference and feature selection.

Table 7  Evaluation results based on TDCCM under different parameter settings on the debutanizer column case.

| Parameter settings | $R^2$ | RMSE | MAE |
|---|---|---|---|
| $E = 3, \tau = 1$ | 0.7243±0.0330 | 0.1013±0.0060 | 0.0840±0.0064 |
| $E = 4, \tau = 1$ | 0.7386±0.0253 | 0.0987±0.0047 | 0.0819±0.0049 |
| $E = 5, \tau = 1$ | 0.7359±0.0295 | 0.0992±0.0055 | 0.0823±0.0057 |
| $E = 6, \tau = 1$ | 0.7355±0.0263 | 0.0989±0.0042 | 0.0820±0.0039 |
| $E = 4, \tau = 2$ | 0.7203±0.0133 | 0.0996±0.0026 | 0.0839±0.0019 |
| $E = 4, \tau = 3$ | 0.7191±0.0153 | 0.1049±0.0020 | 0.0851±0.0020 |
| $E = 4, \tau = 4$ | 0.6832±0.0307 | 0.1088±0.0049 | 0.0898±0.0051 |

Table 8  Evaluation results based on TDPCM under different parameter settings on the debutanizer column case.

| Parameter settings | $R^2$ | RMSE | MAE |
|---|---|---|---|
| $E = 3, \tau = 1$ | 0.7017±0.0684 | 0.1049±0.0120 | 0.0902±0.0125 |
| $E = 4, \tau = 1$ | 0.7191±0.0039 | 0.1025±0.0007 | 0.0849±0.0007 |
| $E = 5, \tau = 1$ | 0.6684±0.0342 | 0.1112±0.0056 | 0.0945±0.0040 |
| $E = 6, \tau = 1$ | 0.7094±0.0200 | 0.1032±0.0037 | 0.0845±0.0041 |
| $E = 4, \tau = 2$ | 0.6100±0.0153 | 0.1207±0.0023 | 0.0977±0.0019 |
| $E = 4, \tau = 3$ | -0.0976±0.0069 | 0.2026±0.0006 | 0.1552±0.0008 |
| $E = 4, \tau = 4$ | 0.0255±0.0950 | 0.1907±0.0092 | 0.1487±0.0048 |

*2) Comparisons with Deep Learning Methods*

Deep learning models, especially those based on recurrent and attention mechanisms, are widely recognized as powerful tools in soft sensor modeling. In this subsection, we evaluate three representative architectures: the long short-term memory (LSTM) network, the gated recurrent unit (GRU) and the Transformer. Each model is followed by a fully connected feedforward network for final prediction. Hyper-parameters are optimized using Bayesian optimization. Further training details are provided in Appendix B.

The evaluation results of the deep learning models on the debutanizer column case are summarized in Table 9. It can be seen that deep learning models show lower predictive accuracy and higher uncertainty than the proposed methods, likely due to limited training data that makes them prone to overfitting. In contrast, the proposed causal feature selection methods infer time-delayed causal relationships that reflect process mechanisms, enabling effective generalization in data-scarce scenarios with better stability and interpretability.

Table 9  Evaluation results of deep learning method comparisons on the debutanizer column case.

| Methods | $R^2$ | RMSE | MAE |
| --- | --- | --- | --- |
| TDCCM | 0.7386±0.0253 | 0.0987±0.0047 | 0.0819±0.0049 |
| TDPCM | 0.7191±0.0039 | 0.1025±0.0007 | 0.0849±0.0007 |
| LSTM | 0.4180±0.1043 | 0.1469±0.0126 | 0.1106±0.0120 |
| GRU | 0.2757±0.1347 | 0.1639±0.0147 | 0.1171±0.0134 |
| Transformer | 0.2954±0.1167 | 0.1618±0.0132 | 0.1249±0.0133 |

*3) Model Adaptability Analysis*

To assess the effectiveness of the proposed model under nonlinear regression, we further evaluate all feature selection methods on the debutanizer column case using Gaussian-kernel support vector regression (SVR). The Gaussian-kernel SVR model is implemented in MATLAB with default hyper-parameter settings. This strategy mitigates overfitting risks under limited data conditions and ensures that the focus remains on evaluating feature selection methods. The evaluation results are presented in Table 10. When the linear PLS model is replaced with the nonlinear SVR model, a general decline in predictive performance is observed across all methods. This may stem from the limited sample size and the use of default parameter settings. Notably, this result should not be viewed as a general flaw of nonlinear regression models, but rather reflects a specific case of the SVR model without parameter tuning. The original model without feature selection achieves superior performance to several baseline methods, suggesting that nonlinear regression can partially mitigate the influence of irrelevant features. Despite the overall performance drop, TDCCM remains the method with highest average performance, and TDPCM continues to provide stable performance in the worst-case scenarios. These results indicate that the proposed feature selection methods based on TDCCM and TDPCM remain effective regardless of the regression model used.

Table 10  Evaluation results of model adaptability analysis using SVR on the debutanizer column case.

| Methods | $R^2$ | RMSE | MAE |
| --- | --- | --- | --- |
| PCC [10] | 0.3809±0.0106 (0.3601, 0.3962) | 0.1521±0.0013 (0.1502, 0.1547) | 0.1111±0.0013 (0.1090, 0.1132) |
| CMI [22] | 0.1985±**0.0076** (0.1766, 0.2143) | 0.1731±**0.0008** (0.1714, 0.1754) | 0.1285±0.0015 (0.1252, 0.1317) |
| RF [14] | 0.5888±0.0131 (0.5698, 0.6403) | 0.1240±0.0020 (0.1160, 0.1268) | 0.0980±0.0026 (0.0886, 0.1025) |
| TDGC | 0.2835±0.0231 (0.2294, 0.3057) | 0.1636±0.0026 (0.1611, 0.1697) | 0.1202±0.0018 (0.1182, 0.1289) |
| TDTE | 0.4267±0.0242 (0.3853, 0.4547) | 0.1464±0.0031 (0.1428, 0.1516) | 0.1203±0.0023 (0.1166, 0.1254) |
| TDCCM | **0.6538**±0.0176 (0.6080, **0.6717**) | **0.1137**±0.0028 (**0.1108**, 0.1211) | **0.0909**±0.0022 (**0.0886**, 0.0965) |
| TDPCM | 0.6333±0.0119 (**0.6191**, 0.6587) | 0.1192±0.0018 (0.1155, **0.1208**) | 0.0935±0.0017 (0.0888, **0.0953**) |
| Original | 0.6058±0.0083 | 0.1214±0.0013 | 0.0987±**0.0012** |

|  | (0.5945, 0.6269) | (0.1181, 0.1231) | (0.0959, 0.1003) |

*Note*: The best results are highlighted in bold and the second-best results are underlined. The maximum and minimum values are reported in parentheses, respectively.

*4) Ablation Study on the Causal Influence Delay Constraint*

The proposed feature selection strategy relies on the assumption that historical measurements with time lags shorter than the identified causal influence delay have a limited impact on the KPI. To further investigate the validity of this delay constraint, an ablation study is conducted by reintroducing short-lag measurements preceding the detected causal delays of auxiliary variables. Specifically, for all the auxiliary variables, the candidate lag windows are expanded by progressively adding up to five pre-delay extensions, with other feature selection strategies unchanged.

Figs. 12 and 13 report the RMSE of the validation and test sets obtained by TDCCM and TDPCM, respectively. The observations show that incorporating a small number of pre-delay measurements can indeed improve the performance in the test set, with the most notable improvement occurring when three additional lags are included. However, adding more pre-delay measurements eventually leads to performance degradation in the test set. Notably, this behavior is not reflected in the validation performance. On the validation set, the predictive accuracy continues to increase as more pre-delay measurements are included. This discrepancy suggests that excessive inclusion of pre-delay measurements may lead to overfitting to validation data while providing limited generalization benefits on unseen test samples, particularly when model selection is guided solely by validation performance.

Table 11 presents a comparison between the original results and those predicted by the best-performing models on the validation set within the considered pre-delay windows. It is evident that incorporating pre-delay windows leads to a decline in model performance. Since validation accuracy increases monotonically with larger pre-delay windows, it fails to clearly indicate when pre-delay measurements cease to be beneficial. For this reason, a causal influence delay constraint is strictly enforced as a conservative design choice to improve generalization and model stability under the current validation strategy.

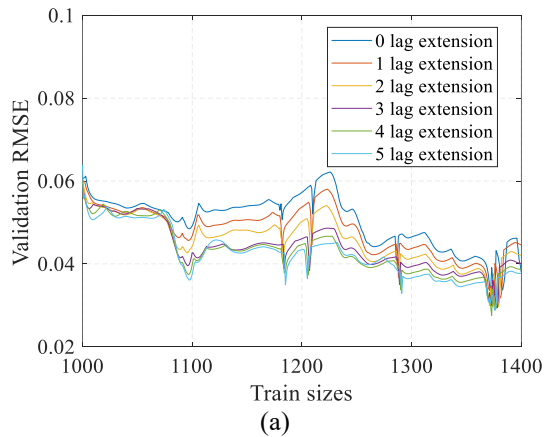

(a)

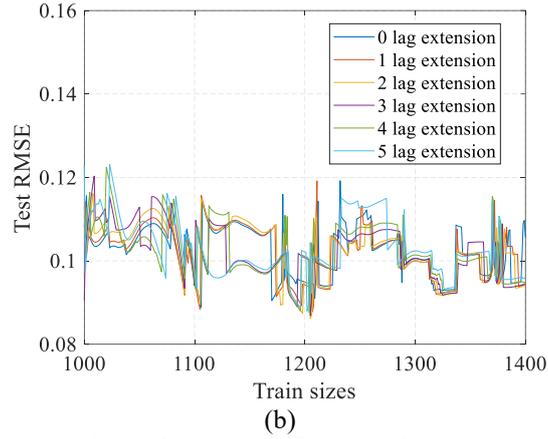

(b)

Fig. 12 RMSE of the ablation study on the causal influence delay constraint calculated by TDCCM for the debutanizer column case: (a) validation set; (b) test set.

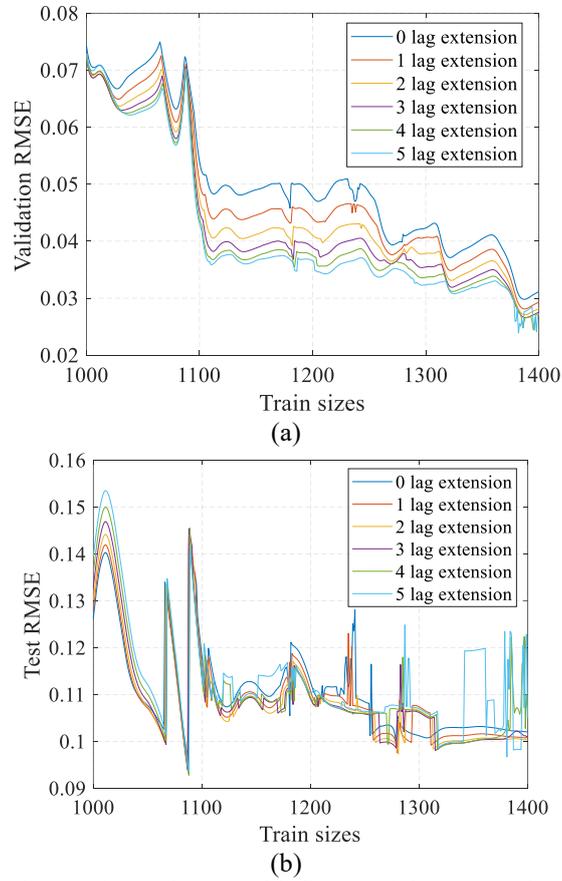

(a)

(b)

Fig. 13 RMSE of the ablation study on the causal influence delay constraint calculated by TDPCM for the debutanizer column case: (a) validation set; (b) test set.

Table 11 Evaluation results of the ablation study on the causal influence delay constraint on the debutanizer column case.

| Methods | $R^2$ | RMSE | MAE |
| --- | --- | --- | --- |
| TDCCM (original) | **0.7386**±0.0253 | **0.0987**±0.0047 | **0.0819**±0.0049 |
| TDPCM (original) | 0.7191±**0.0039** | 0.1025±**0.0007** | 0.0849±**0.0007** |
| TDCCM (lag extension) | 0.7308±0.0223 | 0.1002±0.0041 | 0.0834±0.0036 |
| TDPCM (lag extension) | 0.6853±0.0468 | 0.1082±0.0079 | 0.0912±0.0070 |

*5) Computational Cost Analysis*

In this subsection, the computational cost of different causal inference methods is evaluated. All computations are performed on a laptop equipped with an 11th Gen Intel(R) Core(TM) i7-11800H @ 2.30 GHz CPU and 16 GB RAM. The causal inference considers seven auxiliary variables and one KPI, with a maximum time delay of $d = 100$. For each method, we evaluate the causal strength between every pair of auxiliary variables as well as between each auxiliary variable and the KPI, resulting in 56 variable pairs over 100 time delays. The empirical wall-clock times for the four causal inference methods on this setting are summarized in Table 12. TDGC is computationally efficient, requiring less than half a second to complete the analysis. TDTE is approximately 1,000 times slower. TDCCM and TDPCM incur the highest computational costs, with runtimes around 850 seconds. This aligns with their theoretical complexity, as both involve repeated state-space reconstruction and nearest-neighbor searches. The additional overhead introduced by TDPCM on top of TDCCM is minimal, indicating that the majority of the computational burden stems from the TDCCM step, while the partial correlation calculations in TDPCM contribute only a negligible cost.

In practice, causal inference and feature selection are performed offline and only once for a given process and dataset, whereas the resulting soft sensor model is used repeatedly in online operation. Therefore, although TDCCM and TDPCM are more expensive than TDGC and TDTE, the computational cost of the soft sensor model remains acceptable in industrial applications.

Table 12 Empirical wall-clock times for the four causal inference methods on the debutanizer column case.

| Methods | Runtime (s) |
| --- | --- |
| TDGC | 0.326 |
| TDTE | 176.046 |
| TDCCM | 853.170 |
| TDPCM | 855.241 |

4.2 Continuous Stirred Tank Reactor

4.2.1 Dataset Description

The continuous stirred tank reactor (CSTR) is one of the most widely used reactor types in chemical and biochemical industries. The flowchart of the CSTR is illustrated in Fig. 14, with detailed descriptions of monitoring variables provided in Table 13. Specifically, S1 represents the liquid level in the reactor, which is vital for ensuring stable volume and proper mixing. S2 and S4 denote the coolant flow rate and the coolant temperature in the jacket, respectively. Both of them are essential for temperature regulation within the reactor. S3 is the reactor temperature, a critical variable affecting reaction kinetics. S5 and S7 indicate the feed flow rate and reactant concentration in the reactor feed stream, respectively. Both of them directly influence the input of reactants and the reaction rate. S8 is the feed temperature, which can impact the initial energy state of the reactants entering the reactor. S6 represents the outlet flow rate, helping to maintain a steady-state operation by balancing inflow and outflow. S9, the inlet coolant temperature, is necessary for managing the heat exchange dynamics.

Fig. 14 Flowchart of the CSTR.

Table 13 Monitoring variables on the CSTR case.

| Monitoring variables | Variable description | Unit |
|---|---|---|
| S1 | Liquid level in the reactor | m |
| S2 | Coolant flow rate | L/min |
| S3 | Coolant temperature in the cooling jacket | K |
| S4 | Reactor temperature | K |
| S5 | Feed flow rate of the reactor feed stream | L/min |
| S6 | Outlet flow rate of the reactor | L/min |
| S7 | Reactant concentration in the reactor feed stream | mol/L |
| S8 | Reactor feed temperature | K |
| S9 | Inlet coolant temperature | K |

In a CSTR, reactants are continuously fed into the reactor while products are simultaneously removed, maintaining a constant reaction volume. In practical applications, the reactant concentration in the reactor is the KPI of interest for guiding process control, denoted as S10. To predict the KPI online, the above nine easily measurable auxiliary variables are recorded, denoted as S1-S9. Further details on CSTR can be found in [47]. Based on the process knowledge, the true causal network of the CSTR system is illustrated in Fig. 15 [48].

Fig. 15 True causal network of the CSTR system.

In this study, a total of 2400 samples are collected from the process. The first 1600 samples are used for causal inference and model establishment, and the remaining 800 samples are reserved for model testing. Before analysis, all variables are scaled using min-max normalization.

4.2.2 Causal Inference

Fig. 16 shows the variation of the FNN with the embedding dimension for all the variables in the CSTR case. An embedding dimension of $E = 7$ is then selected, as it is the first dimension at which all FNN values drop below 5%.

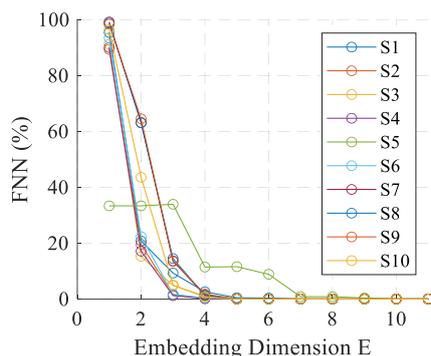

Fig. 16 Variation of FNN with increasing embedding dimension for different variables on the CSTR case.

Then, according to Algorithm 1, the causal inference results of TDPCM can be obtained. A negative lag search window of 20 is employed to identify potential synchrony-induced false causality. Fig. 17 presents the causal strength with time delay of each auxiliary variable on the KPI as calculated by TDCCM and TDPCM. It indicates that S7 has the strongest causal impacts on the KPI, followed by S6 and S1. This result can be explained by process knowledge: S7 determines the amount of reactant entering the reactor, and thus has the most immediate and significant impact on the KPI, as supported by the material balance of the CSTR. S1 influences the KPI by affecting the reaction volume. S6 is inversely related to the residence time of the reactant in the reactor. Theoretically, S5 is expected to have a significant direct causal impact on the KPI. However, because its value is tightly constrained within the process, its influence is not detected. Notably, while S2, S3 and S4 exhibit strong causal intensity with respect to the KPI at early time delays, the extreme values appear at negative time lags, suggesting that these are actually synchrony-induced false causality and the KPI has causal influence on them. This is consistent with the true causal network shown in Fig. 15. The above analysis demonstrates that the causal inference results of TDPCM show satisfactory consistency with the process knowledge.

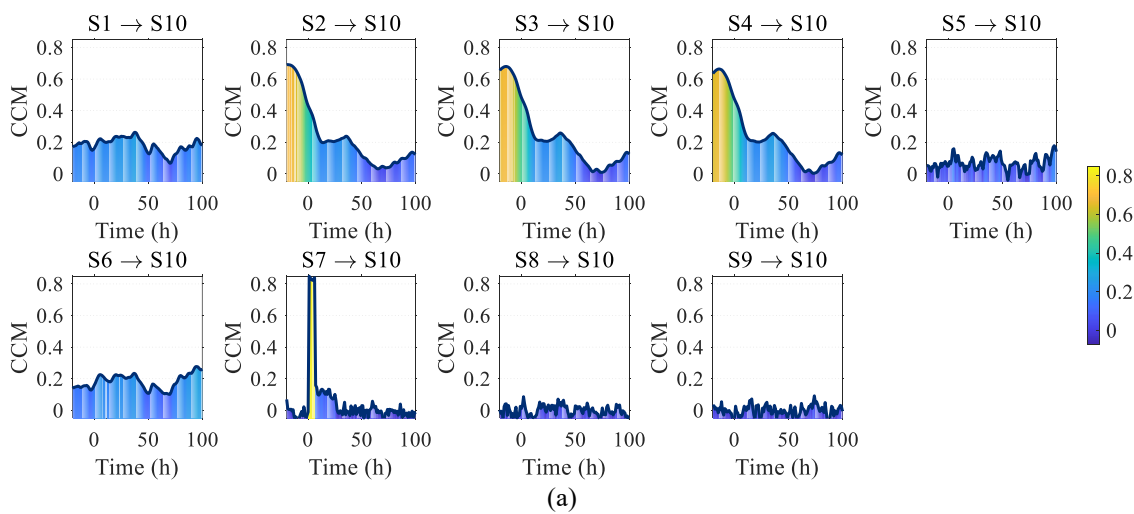

(a)

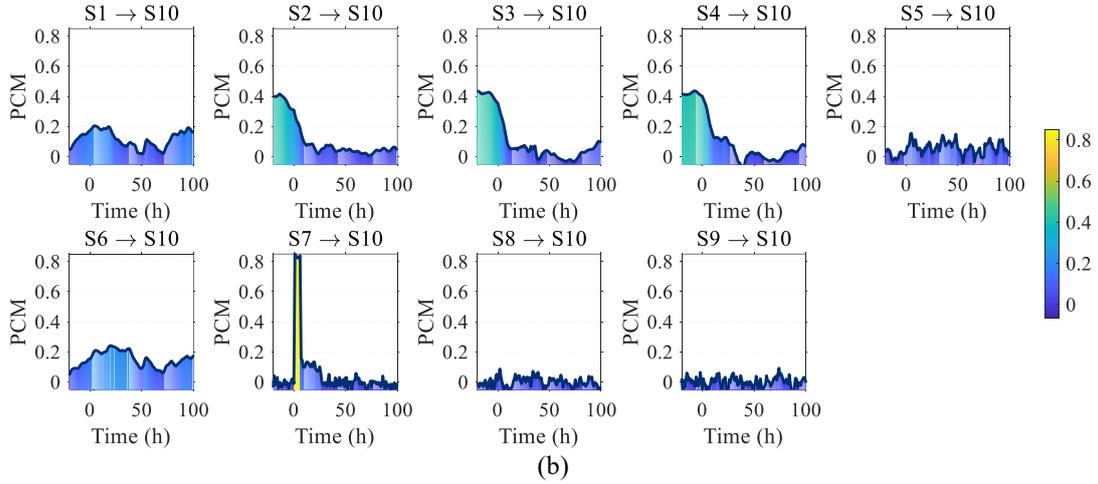

Fig. 17 Causal strength of each auxiliary variable on the KPI on the CSTR case: (a) TDCCM; (b) TDPCM.

4.2.3 Soft Sensor Modeling Results and Comparisons

With a maximum time delay $d$ of 100, 1500 effective labeled samples are available for model development. The sizes of the training set are varied from 1000 to 1350 samples in steps of 5, keeping the sum of training and validation sets constant at 1500, to evaluate the stability of different methods.

Taking a training set sample size of 1250 as an example, Fig. 18 highlights the selected features with the optimized threshold obtained by TDCCM and TDPCM. Notably, the causal influence delays of S2, S3 and S4 are around 25 because the high causal strength at early time delays are synchrony-induced false causality. Therefore, features at these time delays with high causal strength are not selected. In contrast, the causal influence delay for S7 is 1, and its features are selected at early time delay.

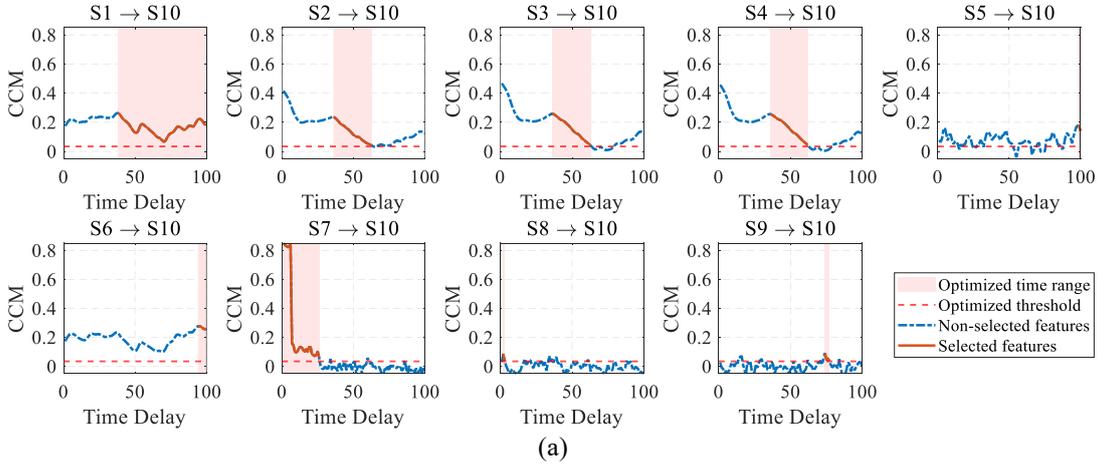

(a)

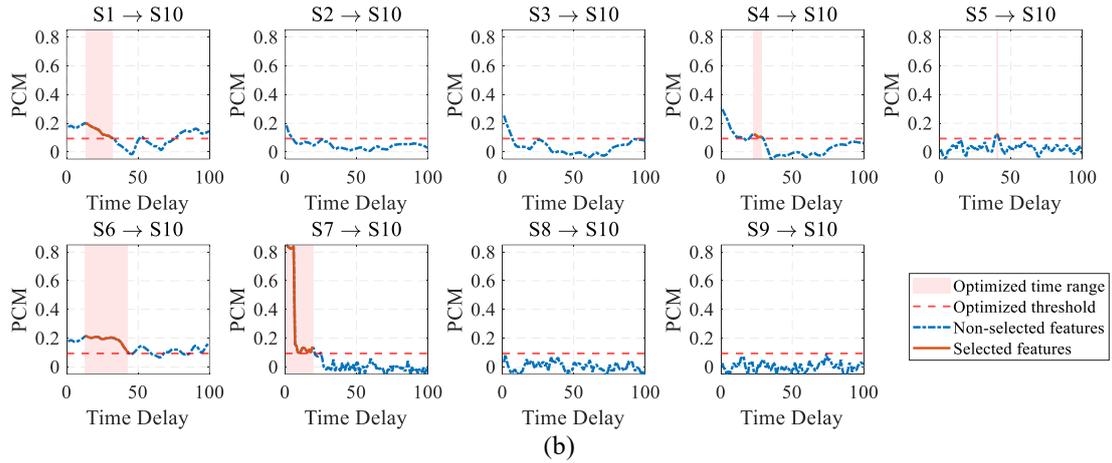

Fig. 18 Selected features of each auxiliary variable with optimized threshold on the CSTR case: (a) TDCCM; (b) TDPCM.

The performance metrics on the test set with increasing sample size of the training set using different feature selection methods are shown in Fig. 19. It can be seen that models based on RF and TDPCM exhibit greater stability compared to other feature selection methods.

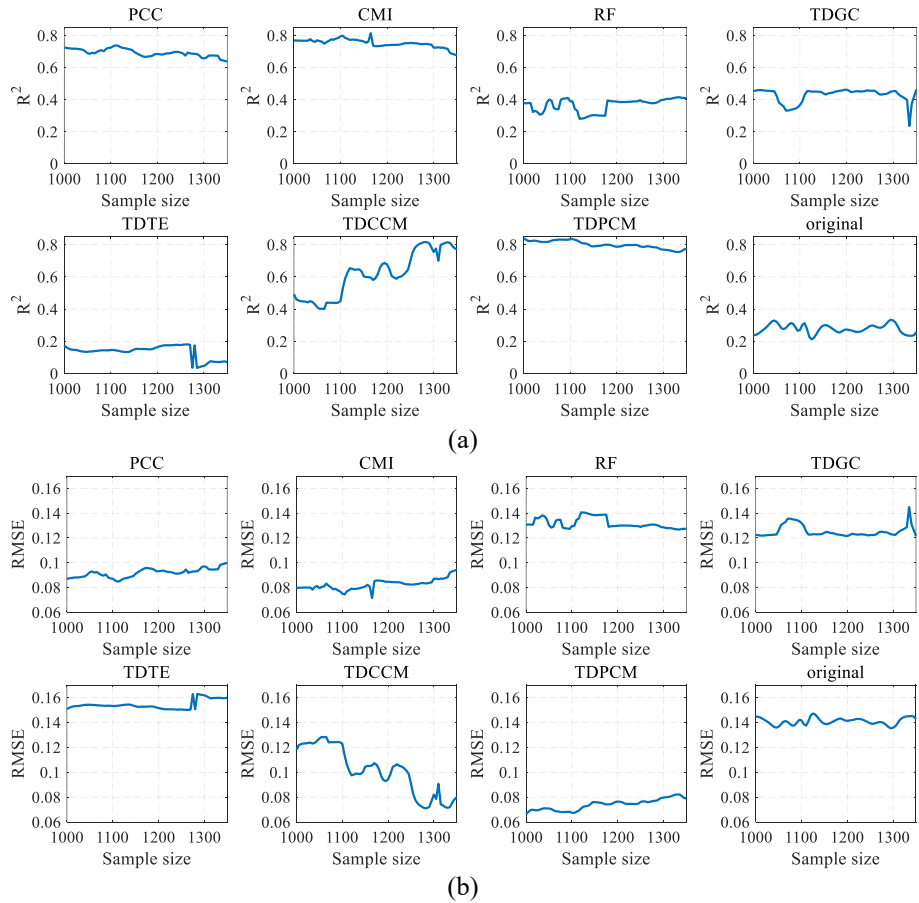

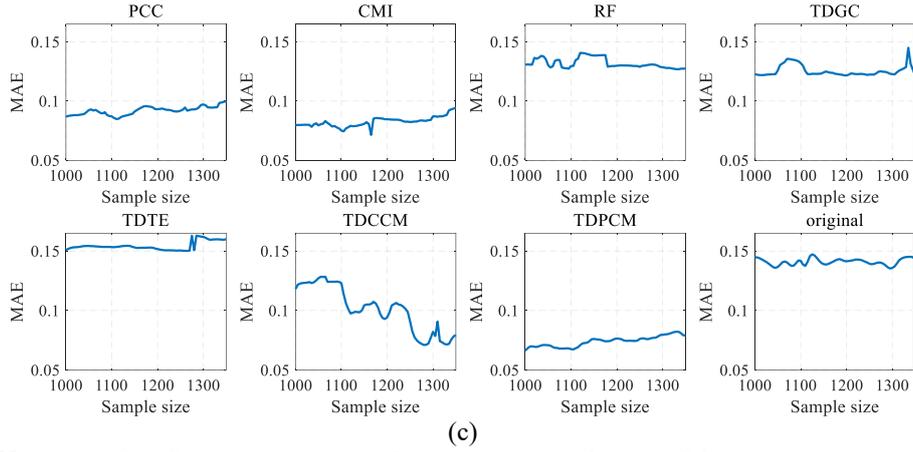

(c)

Fig. 19 Variation of performance metrics with increasing sample size of the training set using different feature selection methods on the CSTR case: (a) $R^2$ (b) RMSE (c) MAE.

Table 14 presents the performance and uncertainties on the test set using different feature selection methods when the sample size of the training set exceeds 1250. For model average performance, TDCCM achieves the best average performance, followed by TDPCM. Compared with the existing best method CMI, TDCCM yields a 7.92% increase in $R^2$, along with 11.33% and 10.04% reductions in RMSE and MAE, respectively. For model uncertainty, the standard deviation of RF is the best, followed by TDPCM. As the method with minimal risk, TDPCM achieves a 7.8% increase in $R^2$ and reductions of 9.46% in RMSE and 13.51% in MAE compared to the second-best method TDCCM in the worst scenario. Overall, the feature selection method based on time-delayed cross mapping techniques achieve superior average performance and stability compared to existing feature selection methods.

Table 14 Evaluation results of different feature selection methods on the CSTR case.

| Methods | $R^2$ | RMSE | MAE |
| --- | --- | --- | --- |
| PCC [10] | 0.6711±0.0172 (0.6378, 0.6933) | 0.0951±0.0025 (0.0919, 0.0998) | 0.0750±0.0025 (0.0716, 0.0794) |
| CMI [22] | 0.7276±0.0250 (0.6771, 0.7538) | 0.0865±0.0039 (0.0823, 0.0943) | 0.0677±0.0034 (0.0638, 0.0743) |
| RF [14] | 0.3975±<u>0.0131</u> (0.3776, 0.4151) | 0.1287±**0.0014** (0.1269, 0.1309) | 0.1038±**0.0012** (0.1024, 0.1056) |
| TDGC | 0.4250±0.0486 (0.2359, 0.4624) | 0.1257±0.0050 (0.1216, 0.1450) | 0.0989±0.0043 (0.0958, 0.1153) |
| TDTE | 0.0949±0.0560 (0.0354, 0.1812) | 0.1577±0.0049 (0.1501, 0.1629) | 0.1293±0.0034 (0.1240, 0.1327) |
| TDCCM | **0.7852**±0.0342 (<u>0.7000</u>, **0.8163**) | **0.0767**±0.0058 (**0.0711**, <u>0.0908</u>) | **0.0609**±0.0051 (**0.0561**, <u>0.0741</u>) |
| TDPCM | <u>0.7735</u>±**0.0123** (**0.7546**, <u>0.7902</u>) | <u>0.0789</u>±0.0021 (<u>0.0760</u>, **0.0822**) | <u>0.0614</u>±0.0017 (<u>0.0590</u>, **0.0641**) |
| Original | 0.2813±0.0329 (0.2344, 0.3347) | 0.1406±0.0032 (0.1353, 0.1451) | 0.1116±0.0027 (0.1071, 0.1153) |

*Note*: The best results are highlighted in bold and the second-best results are underlined. The maximum and minimum values are reported in parentheses, respectively.

Again, we employ the Wilcoxon signed-rank test [46] for significance testing, and the median difference is reported to quantify the practical magnitude of the performance differences. The analysis focuses on the regime where the training sample size exceeds 1250, resulting in 21 paired samples. The corresponding results for TDCCM and TDPCM are summarized in Tables 15 and 16. Across all performance metrics, existing feature selection methods differ significantly from TDCCM and TDPCM, highlighting the consistent superiority of our proposed approach.

Table 15 Wilcoxon signed-rank test and median difference results of TDCCM on the CSTR case.

| Metric | Comparison | $R^+$ | $R^-$ | p-value | Sig. (p<0.001) | Median Δ |
|---|---|---|---|---|---|---|
| $R^2$ | TDCCM vs PCC | 231 | 0 | 3.21E-05 | + | 0.1256 |
| $R^2$ | TDCCM vs CMI | 221 | 10 | 0.000131 | + | 0.0661 |
| $R^2$ | TDCCM vs RF | 231 | 0 | 3.21E-05 | + | 0.3958 |
| $R^2$ | TDCCM vs TDGC | 231 | 0 | 3.21E-05 | + | 0.3690 |
| $R^2$ | TDCCM vs TDTE | 231 | 0 | 3.21E-05 | + | 0.7174 |
| $R^2$ | TDCCM vs TDPCM | 166 | 65 | 0.04112 | / | 0.0160 |
| $R^2$ | TDCCM vs Original | 231 | 0 | 3.21E-05 | + | 0.5128 |
| RMSE | TDCCM vs PCC | 0 | 231 | 3.21E-05 | + | 0.0204 |
| RMSE | TDCCM vs CMI | 9 | 222 | 0.000115 | + | 0.0117 |
| RMSE | TDCCM vs RF | 0 | 231 | 3.21E-05 | + | 0.0538 |
| RMSE | TDCCM vs TDGC | 0 | 231 | 3.21E-05 | + | 0.0515 |
| RMSE | TDCCM vs TDTE | 0 | 231 | 3.21E-05 | + | 0.0825 |
| RMSE | TDCCM vs TDPCM | 61 | 170 | 0.03027 | / | 0.0030 |
| RMSE | TDCCM vs Original | 0 | 231 | 3.21E-05 | + | 0.0665 |
| MAE | TDCCM vs PCC | 1 | 230 | 3.71E-05 | + | 0.0154 |
| MAE | TDCCM vs CMI | 19 | 212 | 0.000424 | + | 0.0081 |
| MAE | TDCCM vs RF | 0 | 231 | 3.21E-05 | + | 0.0445 |
| MAE | TDCCM vs TDGC | 0 | 231 | 3.21E-05 | + | 0.0398 |
| MAE | TDCCM vs TDTE | 0 | 231 | 3.21E-05 | + | 0.0696 |
| MAE | TDCCM vs TDPCM | 86 | 145 | 0.1567 | / | 0.0011 |
| MAE | TDCCM vs Original | 0 | 231 | 3.21E-05 | + | 0.0522 |

Table 16 Wilcoxon signed-rank test and median difference results of TDPCM on the CSTR case.

| Metric | Comparison | $R^+$ | $R^-$ | p-value | Sig. (p<0.001) | Median Δ |
|---|---|---|---|---|---|---|
| $R^2$ | TDPCM vs PCC | 231 | 0 | 3.21E-05 | + | 0.0998 |
| $R^2$ | TDPCM vs CMI | 231 | 0 | 3.21E-05 | + | 0.0392 |
| $R^2$ | TDPCM vs RF | 231 | 0 | 3.21E-05 | + | 0.3706 |
| $R^2$ | TDPCM vs TDGC | 231 | 0 | 3.21E-05 | + | 0.3428 |
| $R^2$ | TDPCM vs TDTE | 231 | 0 | 3.21E-05 | + | 0.687 |
| $R^2$ | TDPCM vs Original | 231 | 0 | 3.21E-05 | + | 0.4991 |
| RMSE | TDPCM vs PCC | 0 | 231 | 3.21E-05 | + | 0.0162 |
| RMSE | TDPCM vs CMI | 0 | 231 | 3.21E-05 | + | 0.0066 |
| RMSE | TDPCM vs RF | 0 | 231 | 3.21E-05 | + | 0.049 |
| RMSE | TDPCM vs TDGC | 0 | 231 | 3.21E-05 | + | 0.0458 |
| RMSE | TDPCM vs TDTE | 0 | 231 | 3.21E-05 | + | 0.0786 |
| RMSE | TDPCM vs Original | 0 | 231 | 3.21E-05 | + | 0.0626 |
| MAE | TDPCM vs PCC | 0 | 231 | 3.21E-05 | + | 0.013 |
| MAE | TDPCM vs CMI | 0 | 231 | 3.21E-05 | + | 0.0056 |
| MAE | TDPCM vs RF | 0 | 231 | 3.21E-05 | + | 0.0418 |
| MAE | TDPCM vs TDGC | 0 | 231 | 3.21E-05 | + | 0.0365 |
| MAE | TDPCM vs TDTE | 0 | 231 | 3.21E-05 | + | 0.0676 |
| MAE | TDPCM vs Original | 0 | 231 | 3.21E-05 | + | 0.0509 |

Fig. 20 shows the comparison between actual values and predicted values of soft sensor models constructed with different feature selection methods when the training set sample size is 1250. The results indicate that the approaches based on TDCCM and TDPCM achieve superior accuracy in overall prediction.

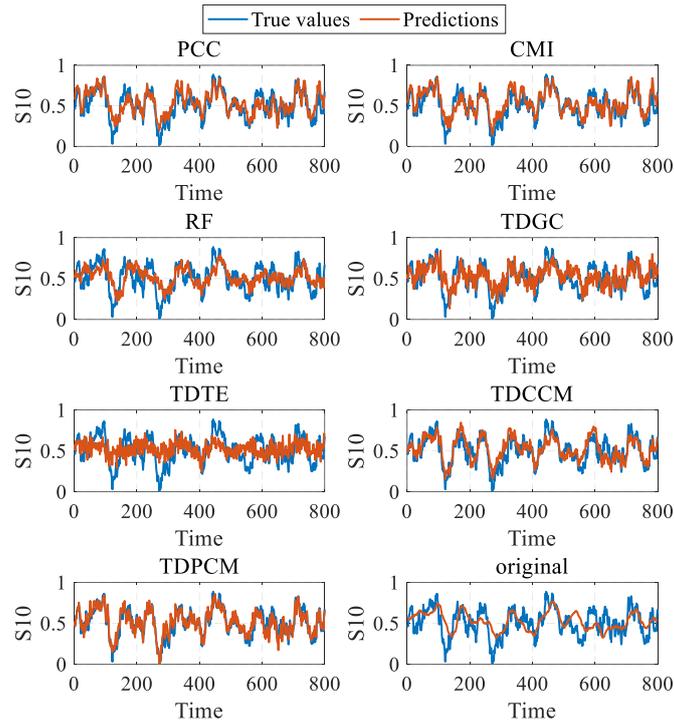

Fig. 20 Comparison between predicted and actual values for each method on the CSTR case.

Table 17 presents the detailed features selected by each method along with the total number of features when the training set sample size is 1250. It can be observed that TDCCM based method has the largest number of features, followed by PCC and TDPCM. However, the accuracy of TDCCM and PCC is much lower than that of TDPCM, indicating that the proposed method enhances model accuracy and stability by reducing redundant features. Moreover, compared with other feature selection methods, although the number of features increases, both model accuracy and stability are improved as listed in Table 14, indicating that important features with direct causality information are involved.

Table 17 Selected features of different methods on the CSTR case.

| Methods | S1 | S2 | S3 | S4 | S5 |
|---|---|---|---|---|---|
| PCC | 1~12 | 1~12 | 1~12 | 1~12 | 1~12 |
| CMI | 1~13 | 1~13 | 1~13 | 1~13 | 1~13 |
| RF | 7, 9, 17, 21, 26, 36, 47, 70, 74 | 1~2, 19, 31, 34~35, 84, 94, 96, 98, 100 | 1~2, 31 | 1~2, 28, 43 | 16 |
| TDGC | 1~4 | 1~15 | 1~14 | 1~11 | 1~3 |
| TDTE | 40~41 | 37~40 | 31~42 | 71~72 | 34 |
| TDCCM | 38~100 | 36~63 | 36~63 | 36~62 | 98~100 |
| TDPCM | 13~32 | / | / | 23~29 | 40~41 |
| Original | 1~100 | 1~100 | 1~100 | 1~100 | 1~100 |
| Methods | S6 | S7 | S8 | S9 | Total number |
| PCC | 1~12 | 1~12 | 1~12 | / | 96 |
| CMI | / | 1~13 | / | / | 78 |
| RF | 6~7, 10~13, 25, 31~32, 35~36, 44, 46, 54 | 1~2 | / | / | 44 |
| TDGC | 1~3 | 1~2 | / | 20 | 53 |
| TDTE | 84~86 | 1~2 | / | 7 | 27 |
| TDCCM | 94~100 | 1~26 | 2~3 | 74~77 | 188 |
| TDPCM | 13~43 | 1~20 | / | / | 80 |
| Original | 1~100 | 1~100 | 1~100 | 1~100 | 900 |

4.2.4 Discussions

*1) Discussions on Causal Inference Results*

In order to show the superiority of the causal inference method based on state-space reconstruction, Fig. 21 depicts the changes in causal strength of S7 on the KPI with different time delays, derived from different time-series causal inference techniques, as well as their optimized thresholds and selected features. Notably, as there are no potential confounding variables in the causal path from S7 to the KPI, TDCCM and TDPCM give similar results.

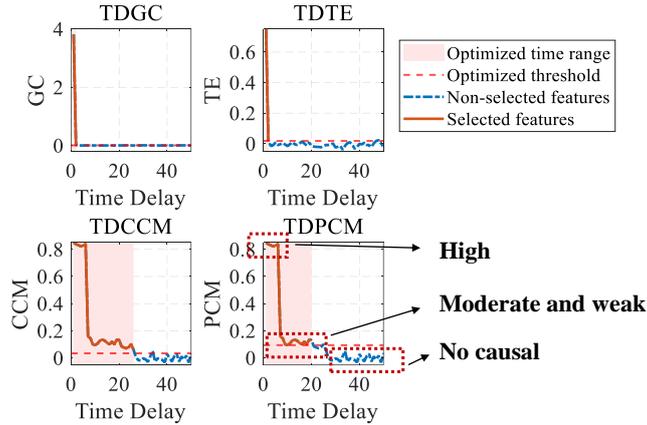

Fig. 21 Variation of causal strengths of S7 on the KPI with various time delays derived by different causal inference methods.

From Fig. 21, both TDGC and TDTE successfully detect the causal relationship from S7 to the KPI, but the causal strength declines sharply as the time delay increases. This occurs because S7 and the KPI are interdependent: the historical information of S7 is already embedded in the past measurements of the KPI. As a result, removing S7 at large time delays has little impact on predicting the KPI. Therefore, for TDGC and TDTE, which rely on the decorrelation assumption, features of S7 with large time delays are considered to have no causal effect. In contrast, methods based on state-space reconstruction infer causal relationships through cross mapping between the manifolds of variables. It can be seen that TDCCM and TDPCM identify three distinct stages of S7 on the KPI: strong causal relationship, moderate or weak causal relationship, and no causal relationship. This allows them to provide more comprehensive information for feature selection compared to TDTE and TDGC.

*2) Ablation Study on the Synchrony-Induced False Causality Identification*

False causality caused by synchrony is a consequence of using state space reconstruction in causal inference [30]. In the proposed causal feature selection framework, TDCCM and TDPCM are designed to explicitly exclude such effects by selecting causal delays only when causal strength is locally maximal. To determine whether this filtering process is necessary, we conduct an ablation experiment where synchrony-induced false causality is intentionally retained. In the CSTR case, variables S2, S3 and S4 are influenced by S10, and the false causality is detected at early time delays. In the ablation variant, these variables are manually assigned causal delays of one, effectively bypassing the false-causality screening. A soft sensor model is then constructed using the resulting feature set, and its predictive performance is compared with that of the original approach.

The evaluation results, summarized in Table 18, show that the inclusion of synchrony-induced false causal features leads to a substantial degradation in model accuracy and stability, with notable increases in RMSE and MAE and a marked reduction in $R^2$. This decline confirms that synchrony-induced false causality does not convey meaningful predictive information and introduce redundant features into the model. Therefore, identifying and removing synchrony-induced false causality is essential for ensuring the effectiveness of causal feature selection results based on time-delayed cross mapping.

Table 18 Evaluation results of the ablation study on the false causality identification on the CSTR case.

| Methods | $R^2$ | RMSE | MAE |
| --- | --- | --- | --- |
| TDCCM (original) | **0.7852**±0.0342 | **0.0767**±0.0058 | **0.0609**±0.0051 |
| TDPCM (original) | 0.7735±**0.0123** | 0.0789±**0.0021** | 0.0614±**0.0017** |
| TDCCM (with false causality) | 0.5462±0.0364 | 0.1117±0.0060 | 0.0876±0.0062 |
| TDPCM (with false causality) | 0.6913±0.0468 | 0.0894±0.0150 | 0.0728±0.0132 |

## 5 Numerical studies

This section aims to assess the ability of TDCCM and TDPCM in inferring causal relationships between interdependent variables considering time delay, which plays a crucial role in effective causal feature selection.

### 5.1 A chain-structured system

First, we consider a chain-structured system to verify whether TDPCM can effectively eliminate indirect causal relationships. Inspired by [32], we establish a four-variable chain-structured system as:

$$
\begin{aligned}
y_1(t) &= y_1(t-1) \cdot [\alpha_1 - \alpha_1 \cdot y_1(t-1)] \\
y_2(t) &= y_2(t-1) \cdot [\alpha_2 - \alpha_2 \cdot y_2(t-1) - \lambda_{12} \cdot y_1(t-1)] \\
y_3(t) &= y_3(t-1) \cdot [\alpha_3 - \alpha_3 \cdot y_2(t-1) - \lambda_{23} \cdot y_2(t-5)] \\
y_4(t) &= y_4(t-1) \cdot [\alpha_4 - \alpha_4 \cdot y_4(t-1) - \lambda_{34} \cdot y_3(t-2)]
\end{aligned}
\tag{26}
$$

where all values of α are set to 3.6; and all values of λ are set to 0.5. To account for the influence of measurement error, Gaussian noise is introduced into each observation, following a normal distribution with zero mean and a standard deviation of 0.01. Its true causal diagram is illustrated in Fig. 22, where the numbers on the links denote the time delays of causal effects.

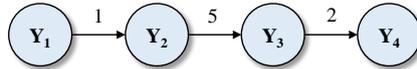

Fig. 22 True causal diagram of the chain-structured system. The numbers on the links denote the time delays of causal effects.

Then, a total of 2000 samples are generated for causal inference. Fig. 23 illustrates the TDCCM and TDPCM results of the chain-structured system. TDCCM infers that $Y_1$, $Y_2$, and $Y_3$ all exert strong causal influences on Y4. In contrast, TDPCM effectively reduces the causal strengths of $Y_1$ and $Y_2$ while preserving that of $Y_3$, demonstrating the capability of TDPCM in identifying direct causality. Additionally, both TDCCM and TDPCM accurately identify the causal delay from $Y_3$ to $Y_4$.

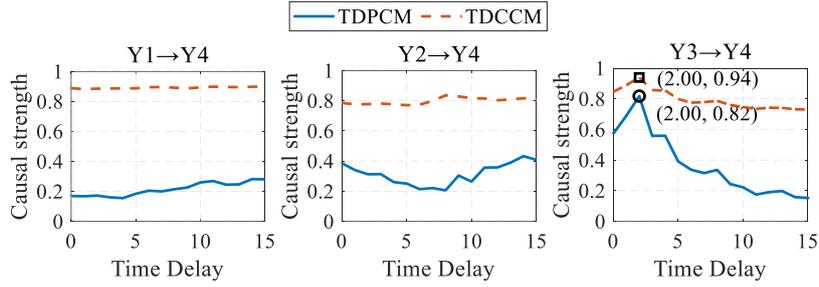

Fig. 23 Causal inference results of TDCCM and TDPCM on the chain-structured system.

We then examine how measurement noise affects the results of causal inference. Specifically, the standard deviation of noise is varied to 0.02, 0.05 and 0.1, while all other parameters remain unchanged. The causal inference results are illustrated in Fig. 24. As noise increases, the inferred causal strength from both methods gradually weakens. At a noise level of 0.02, TDCCM suggests that $Y_1$ has a stronger causal influence than $Y_3$, whereas TDPCM correctly identifies $Y_3$ as having the strongest effect. When the standard deviation increases to 0.05, TDCCM detects an incorrect causal delay, while TDPCM continues to identify the delay accurately. At the highest noise level of 0.1, both methods fail due to excessive noise interference. These results indicate that TDPCM is more robust to noise and more effective at identifying both direct causality and causal delays compared to TDCCM.

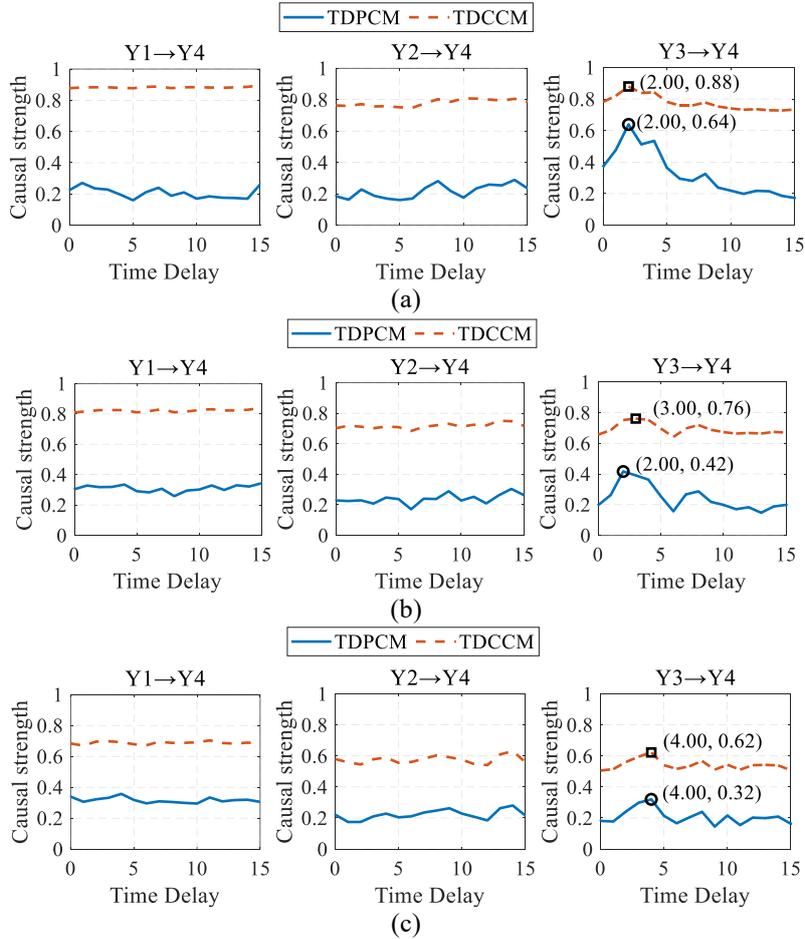

Fig. 24 Causal inference results of TDCCM and TDPCM on the chain-structured system with varying

standard deviation of measurement noise: (a) 0.02; (b) 0.05; (c) 0.1.

5.2 A fork-structured system

A fork-structured system can be used to verify whether TDPCM mistakenly eliminates direct causal relationships. Similarly, we establish a three-variable fork-structured system as:

$$\begin{aligned} y_1(t) &= y_1(t-1) \cdot [\alpha_1 - \alpha_1 \cdot y_1(t-1)] \\ y_2(t) &= y_2(t-1) \cdot [\alpha_2 - \alpha_2 \cdot y_2(t-1)] \\ y_3(t) &= y_3(t-1) \cdot [\alpha_3 - \alpha_3 \cdot y_2(t-1) - \lambda_{13} \cdot y_1(t-1) - \lambda_{23} \cdot y_2(t-3)] \end{aligned} \quad (27)$$

where $\alpha_1 = \alpha_2 = 4$, $\alpha_3 = 2.2$, $\lambda_{13} = 0.6$, $\lambda_{23} = 0.7$. To account for the influence of measurement error, Gaussian noise is again introduced into each observation, following a normal distribution with zero mean and a standard deviation of 0.001. Its true causal diagram is illustrated in Fig. 25, where the numbers on the links denote the time delays of causal effects.

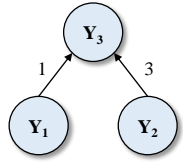

Fig. 25 True causal diagram of the chain-structured system. The numbers on the links denote the time delays of causal effects.

Then, a total of 2000 samples are generated for causal inference. Fig. 26 illustrates the TDCCM and TDPCM results of the fork-structured system. Both TDCCM and TDPCM accurately identify the causal influence and delay from $Y_1$ and $Y_2$ to $Y_3$, indicating that TDPCM does not mistakenly eliminate direct causal relationships.

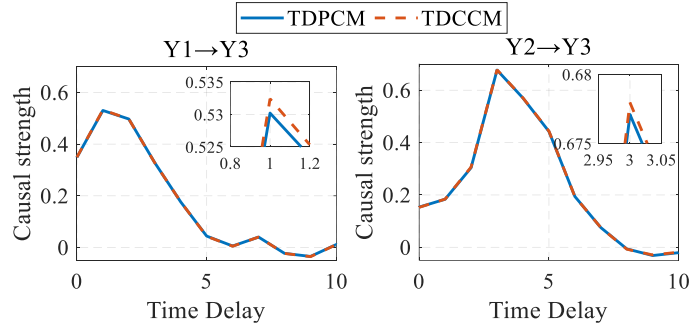

Fig. 26 Causal inference results of TDCCM and TDPCM on the fork-structured system.

## 6 Conclusion

The characteristics of time delays and inherently interdependent variables in industrial processes are always ignored by existing causal feature selection methods, resulting in inadequate model accuracy and stability. To overcome these limitations, this paper proposes a causal feature selection framework based on time-delayed cross mapping. TDCCM is introduced for total causal inference, and TDPCM is developed for direct causal inference. The variation of causal strengths with time delays is considered, and the decorrelation assumption is avoided. Besides, an objective feature selection strategy based on causal inference results is presented. The findings on the industrial cases and numerical studies demonstrate that:

- Compared to causal feature selection method that ignores the influence of varying time delays, TDCCM achieves average RMSE reductions of 34.3% and 11.3% in two respective cases. When compared to causal feature selection methods that overlook the interdependence between variables, TDCCM reduces average RMSE by 22.3% and 38.9%, respectively.
- Compared with TDCCM, TDPCM shows a slight decrease in average performance but exhibits substantially improved stability. As the method with the lowest risk, TDPCM reduces RMSE by 5.93% and 9.46% in two respective cases compared with TDCCM in the worst scenario, , with the average RMSE increasing by only 4.84% and 2.87% in the two cases, respectively.
- For both the chain-structured and fork-structured systems in the numerical studies, TDPCM accurately identifies the direct causal relationships with their causal delays.

Moreover, the proposed method faces several limitations that warrant further research. Firstly, TDPCM relies on the partial Pearson correlation coefficient, which limits its ability to capture nonlinear dependencies. Although some studies have proposed enhancements to CCM to address this shortcoming [49], effectively incorporating nonlinear relationships into the identification of direct causal links remains an open challenge. Secondly, the two case studies considered in this work are systems with relatively smooth and stationary dynamics. In such settings, the causal inference analysis is conducted offline and remains fixed during prediction. However, for processes with regime shifts, multimodality or strong non-stationarity, the causal structure and associated time delays may change over time. A systematic investigation into online extensions therefore represents a promising direction for future research. Finally, this study proposes a feature selection method specifically designed for deterministic regression models. An important future direction involves integrating time-delayed cross mapping techniques with models characterized by inherent uncertainty, such as deep neural networks. This integration may improve key feature identification and ultimately enhance model performance.

## Acknowledgements

This work was supported by the China Scholarship Council [grant number 202406020189], and the National Natural Science Foundation of China [grant number 51775020].

## Appendix A

To ensure reproducibility of the experimental results reported in this study, this appendix summarizes the implementation details and hyper-parameter configurations of all feature selection methods considered. Table 19 lists the calculation methods as well as the specific parameter settings for all the feature selection methods used in the two industrial cases.

Table 19 Implementation details of feature selection methods in the two case studies.

| Methods | Calculation method | Debutanizer column case | CSTR case |
| --- | --- | --- | --- |
| PCC | / | / | / |
| CMI | K-nearest-neighbor | Neighborhood size = 3 | Neighborhood size = 3 |
| RF | / | NumTrees = 100, MinLeafSize = 1, | NumTrees = 100, MinLeafSize = 1, |

| | | SplitCriterion = mse, random seed = 1107 | SplitCriterion = mse, random seed = 1107 |
|---|---|---|---|
| TDGC | Vector autoregression | Model order = 3 | Model order = 3 |
| TDTE | K-nearest-neighbor | Neighborhood size = 3 | Neighborhood size = 3 |
| TDCCM | / | $E = 4, \tau = 1$ | $E = 7, \tau = 1$ |
| TDPCM | / | $E = 4, \tau = 1$ | $E = 7, \tau = 1$ |
| Original | / | / | / |

Appendix B

This appendix summarizes the deep learning architectures and hyper-parameter configurations used in the LSTM, GRU and Transformer models. Hyper-parameters are optimized via Bayesian optimization using Optuna [50]. The complete configuration after optimization is provided in Table 17. All models are trained using mean squared error loss and the Adam optimizer. Data are standardized before model training using min-max normalization. The models are trained ten times on every training set to get a robust result.

Table 20 Neural network architectures and hyper-parameter configurations used in the deep learning models.

| Component | Transformer | GRU | LSTM |
|---|---|---|---|
| Sequence length | 100 | 100 | 100 |
| Batch size | 16 | 16 | 32 |
| Learning rate | $1.22 \times 10^{-3}$ | $3.28 \times 10^{-4}$ | $1.70 \times 10^{-4}$ |
| Dropout probability | 0.4425 | 0.4358 | 0.0819 |
| Hidden size | 32 | 128 | 128 |
| Number of layers | 2 encoder layers | 1 GRU layer | 3 LSTM layers |
| Attention heads | 1 | / | / |
| Feed forward dimension | 512 | / | / |
| Fully connected regressor | 2-layer (64, 64) | 2-layer (32, 16) | 2-layer (32, 128) |
| Training epochs | 200 | 200 | 200 |

46